\documentclass[letterpaper]{article}
\usepackage[preprint]{actlim-arxiv}
\usepackage[hyphens]{url}
\usepackage{graphicx}
\usepackage{natbib}
\usepackage{caption}
\usepackage{amsmath,amssymb,booktabs,multirow,array}
\usepackage{xcolor,colortbl,placeins}
\usepackage[hidelinks,bookmarksnumbered]{hyperref}
\newcommand{\RWTS}{1.0,\,0.3,\,0.2,\,0.1}

\definecolor{lvspec}{HTML}{1B3A6B}
\definecolor{lvfacet}{HTML}{4A7FC1}
\definecolor{lvcat}{HTML}{C8DCF0}
\definecolor{lvnone}{HTML}{D9D9D9}
\newcommand{\badge}[3]{\colorbox{#1}{\textcolor{#2}{\scriptsize\textsf{#3}}}}
\newcommand{\mspec}{\badge{lvspec}{white}{SPECIFIC~1.0}}
\newcommand{\mfacet}{\badge{lvfacet}{white}{FACET~0.6}}
\newcommand{\mcat}{\badge{lvcat}{black}{CATEGORY~0.2}}
\newcommand{\mnone}{\badge{lvnone}{black}{NONE~0.0}}
\newcommand{\key}[1]{\textbf{#1}}
\newcommand{\ours}{\textbf{ActLim-8B}}

\title{Where Research Ideation Begins: Forward-Citation Supervision for Actionable Limitation Generation}
\author{Liyang Zhao\textsuperscript{1}\quad Yong Cao\textsuperscript{2}\quad Haibin Ling\textsuperscript{1}}
\affiliations{
\textsuperscript{1}Westlake University\quad
\textsuperscript{2}University of T\"ubingen\\
\href{mailto:zhaoliyang@westlake.edu.cn}{\texttt{zhaoliyang@westlake.edu.cn}}\quad
\href{mailto:yong.cao@uni-tuebingen.de}{\texttt{yong.cao@uni-tuebingen.de}}\\
\href{mailto:linghaibin@westlake.edu.cn}{\texttt{linghaibin@westlake.edu.cn}}
}
\hypersetup{
pdftitle={Where Research Ideation Begins: Forward-Citation Supervision for Actionable Limitation Generation},
pdfauthor={Liyang Zhao, Yong Cao, Haibin Ling},
pdflang={en-US}
}

\begin{document}
\maketitle
\thispagestyle{plain}

\begin{abstract}
Automated research ideation has advanced rapidly, with multi-agent workflows that generate whole research ideas end to end. These frameworks leave largely unexamined the step that seeds a good idea: identifying an actionable limitation from literature. In this paper, we focus on this first step in the research ideation pipeline. Starting from a pilot study, we find that as large language models (LLMs) scale from 0.6B to 32B parameters, they become far better at recognizing which limitation later work addressed, but no better at generating one, suggesting that the bottleneck lies in the supervision signal rather than model capacity. We therefore construct \textsc{ActLim-Bench} from forward citations, yielding 18,279 evidence-linked supervision triples with human verification. On this signal, we propose a three-stage post-training pipeline: supervised fine-tuning, verifier-filtered rejection-sampling fine-tuning, and reinforcement learning with the verifier as an online dense reward. Experimental results show that our method achieves the highest recall score compared to existing methods among all frontier models. We further show that our generated limitations are preferred in 64.9\% of non-tied human comparisons and improve downstream ideation systems. We will release our data and code upon publication.
\end{abstract}

\section{Introduction}
\label{sec:intro}

The rapid progress of LLMs has sparked growing interest in automated research ideation. Existing multi-agent frameworks can autonomously generate end-to-end research ideas, from literature analysis to proposal generation \citep{wang2024scimon,baek2025researchagent,lu2024aiscientist}. Despite these advances, one fundamental step remains largely overlooked: identifying limitations from existing literature that can serve as the foundation for novel research ideas.

\begin{figure}[t]
\centering
\includegraphics[width=\linewidth]{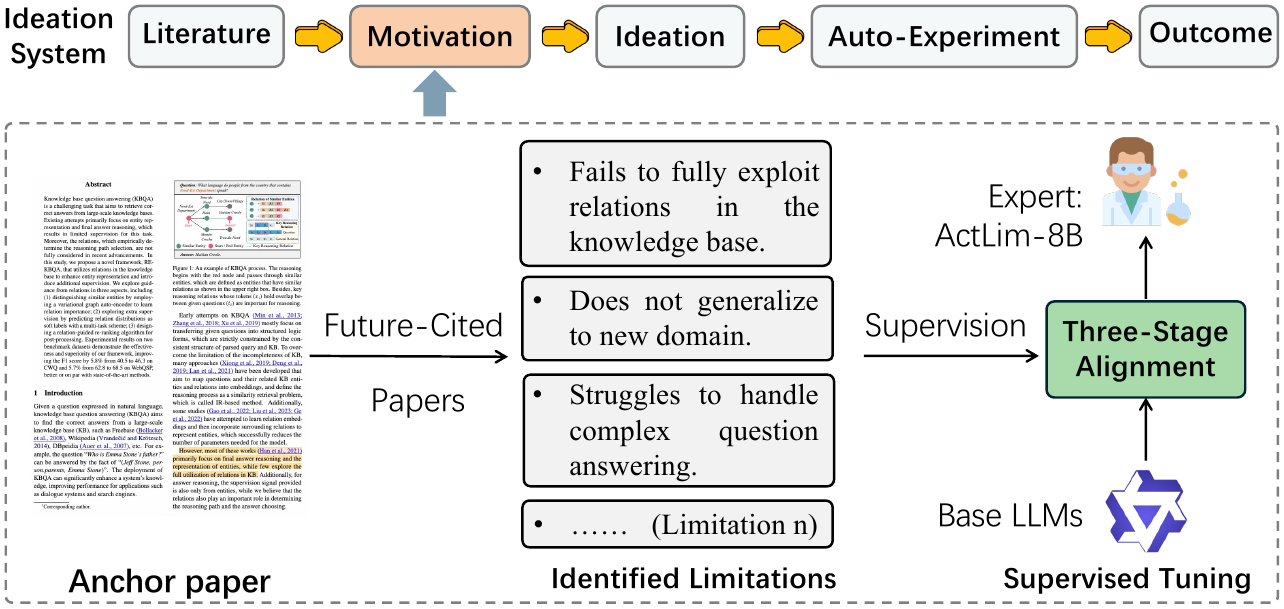}
\caption{Overview of our framework. Using forward-citation supervision as
the training signal, we align general LLMs into domain experts for
actionable limitation generation.}
\label{fig:overview}
\end{figure}

Therefore, as illustrated at the top of Figure~\ref{fig:overview}, we focus on the first stage of the research ideation pipeline: identifying a source-supported limitation of given scientific publication that can motivate concrete follow-up work.
We call this \emph{actionable limitation generation}. The task is narrower than peer review and more inspectable than end-to-end
ideation, reducing error propagation in later stages.

The central challenge is supervision. Existing limitation benchmarks
draw from author-written sections, reviewer critiques, or controlled
synthetic perturbations
\citep{faizullah2024limgen,alazher2025bagels,xu2025limitgen}.
LimAgents adds citation context but still derives reference limitations
from author statements and OpenReview comments
\citep{alazher2026limagents}. Concurrent work uses future publications
to evaluate complete proposals or models idea evolution over citation
graphs
\citep{jiang2026hindsight,wang2026futurealigned,
gao2026graphsofresearch}.

These data sources and attempts capture what authors disclose or reviewers criticize, but
provide limited evidence about which shortcomings later become productive
starting points for research. Nor does model scale resolve this gap. In a
controlled study, larger models become better at selecting a
future-pursued limitation from candidates, while source-only generation
does not improve consistently. This judgment--generation asymmetry
suggests that models can increasingly recognize useful shortcomings when
shown them, yet still lack a direct target for producing them.

Future citations provide that target. Later citing papers reveal which
limitations they adopted as motivation and attempted to address. This
turns a subjective judgment into an auditable supervision
signal. As shown in Figure~\ref{fig:pipeline}, we construct \textsc{ActLim-Bench} by linking each
source-supported limitation to the later citing work that takes it up.
The benchmark contains 18,279 evidence-linked instances, with supporting
spans retained from both the source and citing papers.

We then use the same verified supervision signal in three training stages. Verified
limitations first provide direct demonstrations; the verifier then selects
useful model-generated variants; finally, a set-level reward optimizes
match quality, coverage, redundancy, and format across the complete
three-item response. The resulting 8B generator, ActLim-8B, requires only
the source paper at inference time.

Experimental results show that ActLim-8B reaches a graded recall of 0.452 on the temporally held-out NLP test set, compared with 0.262 for the strongest zero-shot baseline, and
is preferred in blind human evaluation. With NLP-only training, it
transfers to Computer Vision at 0.314 versus 0.252. Its limitations
also improve the retrospective alignment of ideas from a fixed proposer
from 0.302 to 0.480.

In summary, our contributions are threefold: (1) We introduce forward-citation supervision for actionable limitation generation and present \textsc{ActLim-Bench}, \textbf{an auditable benchmark} of 18,279 evidence-linked instances with high annotation quality; (2) We develop \textbf{a three-stage training framework} that exploits the same verified signal for supervision, synthetic data filtering, and reward optimization, yielding ActLim-8B, which requires only the source paper at inference time; (3) \textbf{Extensive experiments} demonstrate the effectiveness of our method through both automatic and human evaluations across two domains.



\section{Related Work}
\label{sec:related}

\paragraph{Scientific feedback and limitation generation.}
Automated scientific feedback includes full-review generation and
reviewer-assistance systems
\citep{yuan2022kidreview,zhou2024reliable,zhu2025deepreview,
li2026reviewgraph,biswas2026aaaipilot,idahl2025openreviewer,
zeng2025reviewrl}. These systems assess broad dimensions such as
correctness, novelty, clarity, and significance, whereas limitation
generation isolates paper shortcomings. Literature-review agents instead
synthesize evidence across collections of papers
\citep{ckmas2025,lira2026}. LimGen
learns from author-written limitation sections
\citep{faizullah2024limgen}; BAGELS combines venue limitations and
external reviews with retrieval \citep{alazher2025bagels}; LimitGen
adds a limitation taxonomy, controlled perturbations, and
human-written targets \citep{xu2025limitgen}; and LimAgents uses
author statements, OpenReview comments, and citation context
\citep{alazher2026limagents}. Our target is different: each instance
links a source paper to a source-supported limitation and to later
citing work that adopts the limitation as motivation and attempts to
address it.

\paragraph{Research ideation and future-grounded evaluation.}
Literature-grounded systems generate hypotheses, research problems,
methods, experiments, or complete proposals using retrieval,
scholarly graphs, critique, and multi-agent refinement
\citep{wang2024scimon,yang2024moose,baek2025researchagent,
li2025chainofideas,lu2024aiscientist}. Dedicated benchmarks assess
research-ideation and scientific-discovery capability
\citep{qiu2025aiideabench,liu2025researchbench,lou2025aaar}. Human
studies report strong
proposal-stage judgments for LLM-generated ideas, while
execution-based evaluation shows that such judgments do not
straightforwardly predict completed research outcomes
\citep{si2025researchideas,si2025executiongap}. Recent work uses
held-out future literature to evaluate proposals or supervise
research-idea generation
\citep{jiang2026hindsight,wang2026futurealigned,
gao2026graphsofresearch}. We operate at a finer, independently
inspectable granularity: the specific shortcoming in a given paper that later work takes up as motivation.

\paragraph{Citation analysis and verifier-guided training.}
Citation contexts support citation-intent classification and
citation-oriented related-work annotation
\citep{cohan2019scicite,li2022corwa}. Temporal citation analysis and
early-impact prediction examine how scientific impact develops over
time \citep{arnaout2026impact,zhao2025wordstoworth}. We use forward-citation contexts not only as
scholarly metadata, but as evidence for a structured generation
target. Our later stages build on self-training, reward-ranked
fine-tuning, and online reinforcement learning
\citep{dong2023raft,shao2024deepseekmath,
deepseekai2025r1}. 

\section{ActLim-Bench}
\label{sec:benchmark}

\paragraph{Source and candidate retrieval.}
To ensure fair training and evaluation, Train and Dev sets contain NLP source papers from 2017--2023. The temporally
held-out Test split contains papers from 2024--2025, and for every pair the
recorded publication date of $Q$ is strictly later than that of $P$. A frozen
extractor reads the abstract and introduction of $Q$ and returns candidate
limitations with citing-paper evidence spans; source-paper evidence is then
retrieved from the abstract, introduction, and conclusion of $P$. The extractor
only proposes candidates. Acceptance is decided by the verifier below.

\paragraph{Forward-citation verification.}
A candidate $\hat{\ell}$ is retained only when
$A(P,Q,\hat\ell)\wedge C(P,\hat\ell)\wedge S(P,\hat\ell)\wedge R(Q,\hat\ell)$.
\textbf{Attribution} $A$ requires that $Q$ attribute the shortcoming to $P$ or
to a clearly defined method family containing $P$. \textbf{Consistency} $C$
requires agreement with the method, assumptions, evidence, and stated scope of
$P$. \textbf{Specificity} $S$ requires a concrete component, mechanism,
assumption, data condition, or evaluation constraint rather than a generic call
for more experiments. \textbf{Response} $R$ requires that the main contribution
of $Q$ attempt to address it. Each accepted record keeps
$(P,\hat\ell,e_P,e_Q,Q)$, where $e_P$ supports consistency and $e_Q$ supports
attribution and response. Each accepted record retains both source- and
citing-paper evidence, making individual targets checkable rather than merely
scored. 


\begin{table}[t]
\centering
\small
\setlength{\tabcolsep}{4.0pt}
\begin{tabular}{lrrrr}
\toprule
Split & Years & Papers & Triples & Targets \\
\midrule
Train & 2017--23 & 4,304 & 12,925 & 10,954 \\
Dev   & 2017--23 & 300   & 901    & 763 \\
Test  & 2024--25 & 1,512 & 4,453  & 3,774 \\
\midrule
Total & 2017--25 & 6,116 & 18,279 & 15,491 \\
\bottomrule
\end{tabular}
\caption{Statistics of \textsc{ActLim-Bench}. Triples are evidence-linked $(P,\ell,Q)$
records; targets are paper-level limitations after within-paper semantic
aggregation. The Computer Vision transfer set is held separately.
}
\label{tab:dataset}
\end{table}

\begin{figure*}[t]
\centering
\includegraphics[width=0.97\textwidth]{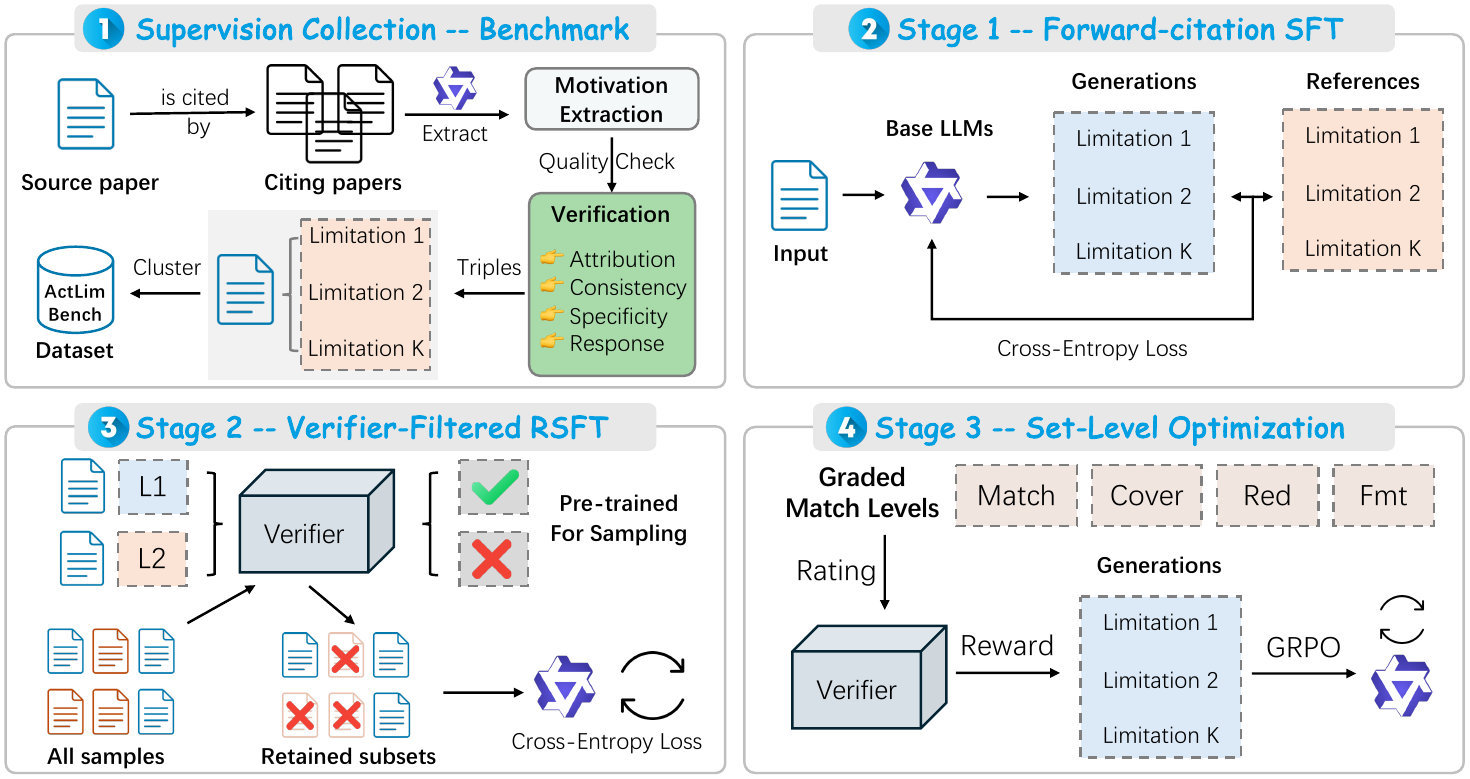}
\caption{\textbf{Benchmark construction and verifier-guided training.}
We construct \textsc{ActLim-Bench} from verified forward citations and
train the model through supervised fine-tuning, verifier-filtered
rejection-sampling fine-tuning, and set-level optimization with graded
rewards.}
\label{fig:pipeline}
\end{figure*}

\paragraph{Aggregation and splits.}
Different citing papers may pursue semantically equivalent limitations of the
same source paper. We merge these into a canonical paper-level target
$\ell\in\mathcal{G}_P$ while retaining every supporting citing paper and span.
We therefore distinguish three units throughout: a citation pair $(P,Q)$, an
evidence-linked triple $(P,\ell,Q)$, and an aggregated target
$\ell\in\mathcal{G}_P$. Splits are built at the source-paper level, with source
and citing papers disjoint across splits. To limit version and near-duplicate leakage we collapse
arXiv and conference versions, normalize metadata, remove highly similar
abstracts, audit citation-graph overlap, and inspect similar limitation
paraphrases across splits. All prompts, extraction rules, and verification
thresholds were fixed before Test was finalized. A separate Computer Vision set
of 2024--2025 papers is built by the identical pipeline and used only for
transfer evaluation. Table~\ref{tab:dataset} summarizes the dataset statistics.

\paragraph{Quality control.}
Two annotators audited 400 accepted triples, equally split between Train and
Test. Requiring every criterion (see Appendix~\ref{app:a-4}) gives a
triple-level precision of 0.87 (95\% Wilson CI $[0.833,0.899]$) and
inter-annotator agreement $\kappa=0.74$ \citep{cohen1960coefficient}. Precision is comparable on Train
(0.88) and Test (0.86). A separate pipeline audit obtains
$\kappa=0.76$ on pair acceptance, and 91\% of accepted pairs contain at
least one fully valid limitation.

Direct attribution accounts for 82\% of the audited triples and has
precision 0.95. The remaining 18\% use attribution to a named method
family containing the source paper, with precision 0.89. Full sampling
strata, rubrics, and adjudicated examples are in the supplement.

\section{Method}
\label{sec:method}

\paragraph{Task definition.}
Let $x_P$ denote the visible text of a source paper $P$, and let
$\mathcal{G}_P$ denote its observed set of future-pursued limitations.
Given $x_P$ and an output number $K$, the model generates
$\mathbf{Y}=(y_1,\ldots,y_K)\sim p_\theta(\cdot\mid x_P,K)$.
Each output should identify a technically specific limitation that is
consistent with the source paper and can motivate concrete follow-up work.
Because $\mathcal{G}_P$ is recovered from later citations, it records
observed future response rather than every valid limitation of $P$.
At inference time, the model receives only $x_P$ and $K$; citing papers,
observed targets, and verifier scores are unavailable to models.

\paragraph{From judgment to supervision.}
Before presenting our method, we first conduct a pilot study to investigate how limitation discrimination and generation scale with model size (see Experiments for more details). The resulting judgment--generation asymmetry (Figure~\ref{fig:scaling}) motivates our training design. When candidate limitations are made explicit, models
increasingly recognize which one later work will pursue, yet this judgment
does not reliably emerge in open-ended generation. Standard post-training
recipes specify how to learn from demonstrations, preferences, or rewards,
but are largely agnostic to which scientific shortcomings should define
those signals. For actionable limitation generation, this target is the
missing ingredient. We therefore use \textbf{forward-citation verification at
three cumulative interfaces (see Figure~\ref{fig:pipeline})}: first to define what the model should imitate,
then to select which of its own generations are worth learning from, and
finally to assign graded credit to a complete response.

\begin{figure}[t]
\centering
\includegraphics[width=\linewidth]{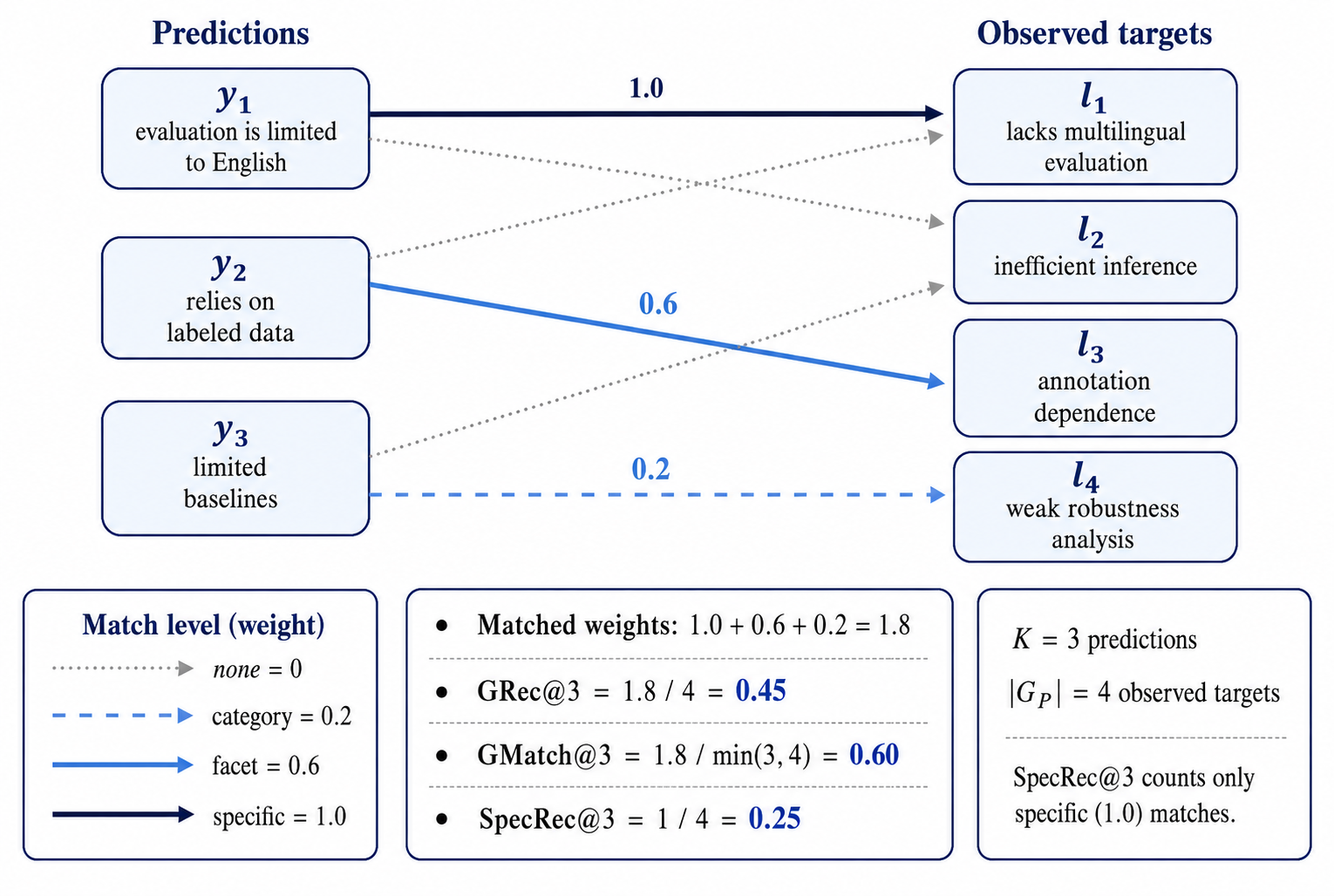}
\caption{\textbf{Graded one-to-one matching.} Predictions are aligned with
observed targets at category, facet, or specific granularity. The
one-to-one constraint prevents broad or repeated predictions from claiming
multiple targets.}
\label{fig:metric}
\end{figure}

\paragraph{Graded verifier.}
The frozen training verifier $V_{\mathrm{train}}$ scores a generated
limitation $y$ against an observed target $\ell$ as
$m_{V_{\mathrm{train}}}(y,\ell)\in\{0,0.2,0.6,1.0\}$, corresponding to
\emph{none}, \emph{category}, \emph{facet}, and \emph{specific} matches.
A category match recovers only the broad type of shortcoming; a facet match
also identifies the affected component and aspect; and a specific match
recovers the concrete shortcoming pursued by later work. The verifier reads
the source paper and the candidate--target pair, but not the citing paper.
Figure~\ref{fig:metric} illustrates the four levels and the one-to-one
matching used throughout training and evaluation process.

\paragraph{Stage 1: Verified demonstrations.}
The first requirement is to make the desired target explicit. The scaling
result shows that models can recognize a future-pursued limitation when
alternatives are provided; Stage 1 turns such limitations into supervised
generation targets. We perform supervised fine-tuning (SFT) on verified
future-pursued limitations. For each paper, we serialize
$N_P=\min(K,|\mathcal{G}_P|)$ targets into a structured response and
explicitly request $N_P$ outputs. When $|\mathcal{G}_P|>K$, we use a fixed
subset selected to maximize diversity in the supporting citing evidence;
the subset is not resampled across epochs. When $|\mathcal{G}_P|<K$, we
request the observed number rather than padding the response with
unsupported items. This count-conditioned objective teaches both the
target construct and the required list structure.

\paragraph{Stage 2: Verifier-guided expansion.}
Verified demonstrations provide precise anchors, but each paper exposes
only a finite set of canonical formulations. The Stage-1 policy can produce
additional paraphrases and refinements, while model likelihood or
reference-free self-selection cannot determine whether these variants
remain aligned with future-pursued targets. We therefore turn the verifier
into a data-selection rule and perform verifier-filtered
rejection-sampling fine-tuning (RSFT). From the Stage-1 policy, we sample a
candidate pool $\mathcal{C}_P$ and retain facet-or-better matches:
\[
\widetilde{\mathcal{C}}_P
=
\left\{
y\in\mathcal{C}_P:
\max_{\ell\in\mathcal{G}_P}
m_{V_{\mathrm{train}}}(y,\ell)\ge 0.6
\right\}.
\]
After removing semantic near-duplicates with a multilingual text embedding
model \citep{chen2024bgem3}, we greedily group retained candidates into
diverse $K$-item responses and continue fine-tuning.
This stage broadens how a verified target can be expressed without
replacing future-grounded selection with the generator's own confidence.

\paragraph{Stage 3: Set-level optimization.}
Candidate filtering uses the verifier as a threshold and evaluates
limitations largely one at a time. The task, however, requires a set of
complementary shortcomings: each item should match deeply, the response
should cover distinct targets, and repeated paraphrases should not receive
additional credit. We therefore retain the verifier's graded judgment and
apply it online to complete responses using GRPO
\citep{shao2024deepseekmath,deepseekai2025r1}. For
$\mathbf{Y}=(y_1,\ldots,y_K)$, the reward is
\[
r(P,\mathbf{Y})
=
\alpha\,\operatorname{Match}
+\beta\,\operatorname{Cover}
-\gamma\,\operatorname{Red}
+\delta\,\operatorname{Fmt},
\]
where
\[
\begin{aligned}
\operatorname{Match}(P,\mathbf{Y})
&=
\frac{1}{|\mathcal{G}_P|}
\max_{\mathcal{M}}
\sum_{(y,\ell)\in\mathcal{M}}
m_{V_{\mathrm{train}}}(y,\ell),\\
\operatorname{Cover}(P,\mathbf{Y})
&=
\frac{
\max_{\mathcal{M}_{0.6}}|\mathcal{M}_{0.6}|
}{
\min(K,|\mathcal{G}_P|)
}.
\end{aligned}
\]
Here $\mathcal{M}$ is a maximum-weight one-to-one matching, and
$\mathcal{M}_{0.6}$ is a maximum-cardinality one-to-one matching over
facet-or-better edges. Match measures graded target recovery, whereas
Cover explicitly rewards distinct facet-or-better matches. $\operatorname{Red}$ is the mean pairwise semantic
similarity among generated limitations, and $\operatorname{Fmt}$ is one
only when the response contains exactly $K$ parseable items. We select
$(\alpha,\beta,\gamma,\delta)=(\RWTS)$ on the Dev set.

Starting from the Stage-2 checkpoint, the current policy samples a group
of complete responses for each source paper. The frozen verifier scores
each response, and GRPO updates the policy using relative advantages within
the group, with a KL penalty to the Stage-2 policy.

\paragraph{One signal, three forms of supervision.}
The verifier retains the same semantics across all three stages while its
training role changes. Stage 1 uses future-pursued limitations as targets;
Stage 2 uses the same judgment to curate model-generated training data; and
Stage 3 uses it to score the quality and coverage of a complete response.
The stages therefore ground \emph{what} to generate, broaden \emph{how}
the target can be expressed, and coordinate multiple limitations as a set.
The verifier is a training-time scaffold: ActLim-8B requires neither
citing papers nor verifier calls at deployment.

\section{Experiments}
\label{sec:experiments}

We evaluate our method via two domains, then isolate the two choices that
define ActLim: the source of supervision and the role assigned to the
verification signal. We next test the judgment--generation asymmetry across
model scale, and close with human and downstream validation.

\subsection{Experimental Setup}

\paragraph{Dataset and baselines.}
We evaluate our method on papers from both NLP and CV domains. Each system receives the abstract, introduction, and conclusion and
returns exactly $K=3$ limitations. Citing papers, evidence
spans, and verified targets are withheld. Retrieval baselines access only
documents published no later than the source paper.

For comparison, we select representative open source models and agentic methods. Qwen3-8B and Qwen3-32B \citep{yang2025qwen3} provide same-family
capability references for the trainable Qwen3-8B generator. Qwen-72B-Chat
\citep{bai2023qwen}, Claude-3-Opus, Gemini-1.0-Ultra
\citep{geminiteam2023gemini}, and GPT-4-Turbo provide additional
zero-shot references. Exact API snapshots, release dates, and documented
knowledge cutoffs are reported in the supplement. Prompting baselines use Qwen3-8B
\citep{wei2022cot,madaan2023selfrefine}. Retrieval and multi-agent
comparisons include pre-publication RAG \citep{lewis2020rag}, a MARG-style
system adapted to three outputs \citep{darcy2024marg}, and BAGELS-RAG with
retrieval restricted to pre-publication documents
\citep{alazher2025bagels}. Trainable conditions use three seeds. Main-table
entries are seed means; paired intervals use 10,000 paper-level bootstrap
resamples after averaging aligned seed scores. Human pairwise tests use exact
binomial tests with Holm correction \citep{holm1979simple}. See supplement for detailed prompts and parameter settings.

\paragraph{Evaluators and metrics.}
The construction evaluator, training verifier $V_{\mathrm{train}}$, and
benchmark evaluator $V_{\mathrm{B}}$ use one frozen DeepSeek model identifier
with role-specific prompts, following LLM-as-a-judge evaluation
\citep{liu2023geval,zheng2023llmjudge}; only the construction evaluator sees
citing papers. The independent evaluator $V_{\mathrm{I}}$ uses GLM-5.2 \citep{glm5team2026glm5} and is not
used for construction, filtering, reward, or model selection. We therefore
treat $V_{\mathrm{I}}$ as primary and report $V_{\mathrm{B}}$ as a
same-benchmark robustness check.

For source paper $P$ and generated set
$\mathbf{Y}_P=(y_1,y_2,y_3)$, let
\begin{equation*}
\begin{aligned}
W_V(P,\mathbf{Y}_P)
&=
\max_{\mathcal{M}}
\sum_{(y,\ell)\in\mathcal{M}}m_V(y,\ell),\\
\operatorname{GRec@3}(P,\mathbf{Y}_P)
&=
\frac{W_V(P,\mathbf{Y}_P)}{|\mathcal{G}_P|},\\
\operatorname{GMatch@3}(P,\mathbf{Y}_P)
&=
\frac{W_V(P,\mathbf{Y}_P)}
{\min(3,|\mathcal{G}_P|)}.
\end{aligned}
\end{equation*}
The maximum ranges over one-to-one matchings
$\mathcal{M}\subseteq\mathbf{Y}_P\times\mathcal{G}_P$ and is solved
by the Hungarian algorithm \citep{kuhn1955hungarian}.
GRec@3 is macro-averaged over papers. SpecRec@3 uses the same matching
with binary weights $\mathbb{I}[m_V(y,\ell)=1]$.
GMatch@3 removes the output-count ceiling when
$|\mathcal{G}_P|>3$ and is used only in the target-count analysis.

\subsection{Main Results}

\begin{table*}[t]
\centering
\small
\resizebox{0.9\textwidth}{!}{%
\setlength{\tabcolsep}{2.8pt}
\begin{tabular}{lccc ccc}
\toprule
\textbf{System}
& \multicolumn{3}{c}{\textbf{NLP Test}}
& \multicolumn{3}{c}{\textbf{CV Transfer}} \\
\cmidrule(lr){2-4}\cmidrule(lr){5-7}
& GRec@3$_{\mathrm{I}}$
& SpecRec@3$_{\mathrm{I}}$
& GRec@3$_{\mathrm{B}}$
& GRec@3$_{\mathrm{I}}$
& SpecRec@3$_{\mathrm{I}}$
& GRec@3$_{\mathrm{B}}$ \\
\midrule
\multicolumn{7}{c}{\textit{Zero-shot}} \\
Qwen3-8B$^\dagger$
& 0.204 & 0.079 & 0.210
& 0.198 & 0.070 & 0.204 \\
Qwen3-32B$^\dagger$
& 0.181 & 0.058 & 0.188
& 0.176 & 0.051 & 0.183 \\
Qwen-72B-Chat
& 0.235 & 0.076 & 0.242
& 0.226 & 0.066 & 0.233 \\
Claude-3-Opus
& 0.251 & 0.089 & 0.258
& 0.243 & 0.078 & 0.250 \\
Gemini-1.0-Ultra
& \underline{0.262} & 0.094 & 0.266
& \underline{0.252} & 0.083 & 0.256 \\
GPT-4-Turbo
& 0.258 & \underline{0.098} & \underline{0.271}
& 0.249 & \underline{0.086} & \underline{0.262} \\
\midrule
\multicolumn{7}{c}{\textit{Prompting and refinement (Qwen3-8B)$^\dagger$}} \\
\quad + CoT
& 0.209 & 0.077 & 0.218
& 0.203 & 0.068 & 0.212 \\
\quad + LimGen-style
& 0.212 & 0.080 & 0.221
& 0.204 & 0.070 & 0.213 \\
\quad + Self-Refine
& 0.215 & 0.081 & 0.226
& 0.209 & 0.071 & 0.220 \\
\midrule
\multicolumn{7}{c}{\textit{Retrieval and multi-agent}} \\
Pre-publication RAG$^\dagger$
& 0.256 & 0.086 & 0.265
& 0.238 & 0.074 & 0.246 \\
MARG-style (adapted)
& 0.233 & 0.066 & 0.238
& 0.219 & 0.058 & 0.224 \\
BAGELS-RAG (pre-pub.)
& 0.240 & 0.084 & 0.244
& 0.223 & 0.073 & 0.227 \\
\midrule
\multicolumn{7}{c}{
\textit{ActLim training stages (three-seed means; Qwen3-8B)$^\dagger$}} \\
Stage 1: Verified SFT
& 0.352 & 0.171 & 0.361
& 0.232 & 0.094 & 0.238 \\
Stage 2: +Verifier-filtered RSFT
& 0.410 & 0.226 & 0.428
& 0.279 & 0.129 & 0.291 \\
Stage 3: ActLim-8B (+set-level GRPO)
& \textbf{0.452} & \textbf{0.241} & \textbf{0.485}
& \textbf{0.314} & \textbf{0.141} & \textbf{0.337} \\
\bottomrule
\end{tabular}}
\caption{\textbf{Main results on the held-out NLP test set and zero-shot
transfer to CV.}
All trainable models use NLP-only training. Independent-evaluator metrics
are primary; GRec@3$_{\mathrm{B}}$ shows that the ranking is preserved under
the benchmark evaluator. Bold marks the overall best result and underline
the strongest zero-shot result in each column.
$^\dagger$ indicates a Qwen3-based system, whose pretraining window may
overlap evaluation papers. Baseline adaptations and more details
are reported in the supplement.}
\label{tab:main}
\end{table*}

\paragraph{Held-out NLP results.}
As shown in Table~\ref{tab:main}, ActLim-8B reaches 0.452
GRec@3$_{\mathrm{I}}$ and 0.241 SpecRec@3$_{\mathrm{I}}$, compared
with the strongest zero-shot values of 0.262 and 0.098. The 8B model therefore gains 0.190 graded recall, more than
doubles specific-level recovery, and exceeds the same-family 32B reference.
The benchmark evaluator gives the same ordering, 0.485 versus 0.271.
The cumulative stages are also visible in the table: verifier-guided expansion
adds 0.058 GRec@3$_{\mathrm{I}}$ and 0.055 SpecRec@3$_{\mathrm{I}}$ over
verified SFT; set-level GRPO adds a further 0.042 and 0.015. Expansion
contributes most of the specific recovery, while the set-level objective
adds graded alignment and coverage across the three-item response.

\paragraph{Transfer to CV.}
The CV test set contains 1,737 source papers from CVPR, ICCV, and ECCV
2024--2025, with 3,961 evidence-linked triples and 3,562 aggregated targets.
No CV data are used for training or model selection. ActLim-8B reaches 0.314
GRec@3$_{\mathrm{I}}$, 0.062 above the strongest zero-shot system
(95\% CI $[0.048,0.076]$), and 0.116 above Qwen3-8B
($[0.102,0.130]$). SpecRec@3$_{\mathrm{I}}$ rises from the strongest
zero-shot score of 0.086 to 0.141. The Stage-1--Stage-2--Stage-3 sequence
continues at 0.232, 0.279, and 0.314, showing that both verifier-guided
expansion and set-level optimization transfer beyond the training domain.
A separate 100-triple audit yields 0.83 joint precision
(95\% Wilson CI $[0.745,0.891]$; $\kappa=0.71$). CV papers contain fewer
observed targets on average than NLP papers (2.05 versus 2.50), so absolute
recall values are interpreted within domain.

\paragraph{Target-count and length controls.}
The gains persist across target-count strata and after matching output
length. ActLim-8B produces shorter outputs than Qwen3-32B
(19.7 versus 29.9 tokens); stratified analyses are in the supplement.

\subsection{Why Forward-Citation Supervision Helps}

To isolate the supervision source, we train four models on the same 1,240
papers, 2,795 examples, and approximately 2.1M target tokens, using the same
recipe and three seeds. Target-length distributions are matched. Forward-
citation supervision reaches 0.353 GRec@3$_{\mathrm{I}}$, 0.117 above the
strongest alternative, and 0.172 SpecRec@3$_{\mathrm{I}}$, compared with
0.087 for controlled perturbations (Table~\ref{tab:source}). The benchmark
evaluator gives a 0.120 gain
(95\% CI $[0.101,0.140]$). The effect is therefore not limited to broad
category agreement; it is largest at the specific level that can support a
concrete follow-up.

The author-supervised model provides a useful diagnostic. It reaches 0.47
GRec@3 on held-out author targets but 0.193 on future-pursued targets. The
model can fit the author-written signal; what changes is which shortcomings
the supervision teaches it to recover.

\begin{table}[t]
\centering
\small
\setlength{\tabcolsep}{3.8pt}
\begin{tabular}{@{}lccc@{}}
\toprule
Supervision
& GRec@3$_{\mathrm{I}}$
& SpecRec@3$_{\mathrm{I}}$
& GRec@3$_{\mathrm{B}}$ \\
\midrule
Author-stated
& 0.193 & 0.072 & 0.198 \\
Reviewer-written
& 0.219 & 0.081 & 0.224 \\
Controlled perturbations
& 0.236 & 0.087 & 0.242 \\ \midrule
Forward-citation
& \textbf{0.353} & \textbf{0.172} & \textbf{0.362} \\
\bottomrule
\end{tabular}
\caption{\textbf{Matched supervision-source comparison.}
All conditions use the same source papers, number of examples, token
budget, target-length distribution, training recipe, and three seeds.
Values are three-seed means; all standard deviations are at most 0.007
and are reported in the supplement.}
\label{tab:source}
\end{table}

\subsection{How Verification Enters Training}

Starting from the same Stage-1 checkpoint, DPO
\citep{rafailov2023dpo}, verifier-guided expansion, and matched set-level
GRPO each consume 9,000 generated items and 9,000 verifier calls. They differ
only in whether the judgment becomes a binary preference, an item-level
selection rule, or a response-level reward. As Table~\ref{tab:integration}
shows, verifier-guided expansion reaches 0.410 GRec@3$_{\mathrm{I}}$, while
the matched set-level objective reaches 0.425. Combining expansion with
set-level GRPO gives the full ActLim-8B result of 0.452. Under the matched
budget, response-level optimization improves over candidate filtering by
0.015 GRec@3$_{\mathrm{I}}$ and 0.021 GRec@3$_{\mathrm{B}}$
(95\% CI $[0.005,0.037]$).

The reward components make complementary contributions. Match alone reaches
0.458 GRec@3$_{\mathrm{B}}$; adding coverage, redundancy, and format terms
raises the score successively to 0.472, 0.481, and 0.485. These results mirror
the three roles developed in the method: verified targets establish what to
generate, expansion broadens acceptable realizations, and set-level reward
coordinates a complete response.

\begin{table}[t]
\centering
\small
\resizebox{\columnwidth}{!}{%
\setlength{\tabcolsep}{1.0pt}
\begin{tabular}{@{}llcc@{}}
\toprule
Strategy & Use of judgment
& GRec@3$_{\mathrm{I}}$ & GRec@3$_{\mathrm{B}}$ \\
\midrule
Self-rerank@10
& None
& 0.371 & 0.390 \\
DPO
& Binary preference
& 0.365 & 0.382 \\
Verifier-guided expansion
& Item filter
& 0.410 & 0.428 \\
Set-level GRPO (matched)
& Response reward
& 0.425 & 0.449 \\
\midrule
Full: expansion + GRPO
& Filter+response reward
& \textbf{0.452} & \textbf{0.485} \\
\bottomrule
\end{tabular}}
\caption{\textbf{Ablation study on Verifier.}
DPO, expansion, and matched set-level GRPO share the Stage-1
initialization, source papers, generated-item and verifier-scoring
volume. Self-rerank@10 is an inference-only reference; the final row combines
the two complementary verifier-guided stages.}
\label{tab:integration}
\end{table}

\subsection{Judgment--Generation Scaling}

We evaluate Qwen3 at 0.6, 1.7, 4, 8, 14, and 32B parameters. Each selection
item contains one future-pursued limitation and seven counterfactual negatives
matched for category, length, specificity, lexical style, and source
relevance; chance is 12.5\%. An ambiguity screen removes 91 of 1,512 sets,
and 97\% of 200 audited retained sets have a unique correct answer. A
candidate-only classifier that does not see the source paper scores
$0.138\,[0.121,0.156]$, with an interval containing chance.

\begin{figure}[t]
\centering
\includegraphics[width=\columnwidth]{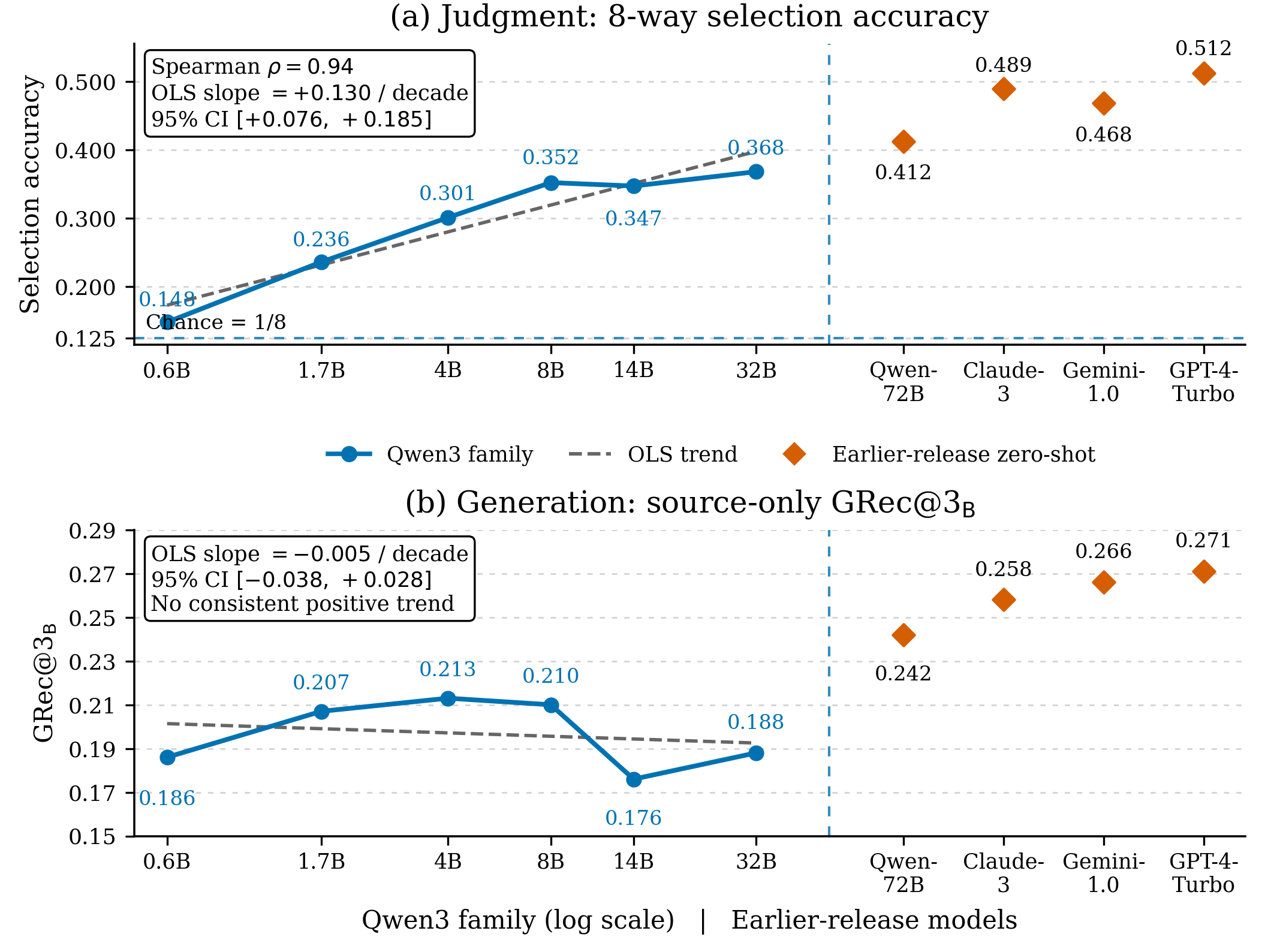}
\caption{\textbf{Judgment--generation asymmetry.}
Within the Qwen3 family, 8-way selection accuracy rises strongly with model
size, whereas source-only limitation generation shows no consistent positive
trend. Earlier-release zero-shot systems are complementary references and are
excluded from the same-family regressions. The horizontal dashed line in
panel (a) marks chance at $1/8$.}
\label{fig:scaling}
\end{figure}

Selection accuracy rises from 0.148 to 0.368 across the Qwen3 family
(Spearman $\rho=0.94$; OLS slope $+0.130$ per decade,
95\% CI $[0.076,0.185]$). Generation on the same papers shows no consistent
positive trend: GRec@3$_{\mathrm{B}}$ ranges from 0.176 to 0.213, with slope
$-0.005$ per decade and 95\% CI $[-0.038,0.028]$. Earlier-release systems
show the same qualitative separation, with selection accuracies of
0.412--0.512 and generation scores of 0.242--0.271
(Figure~\ref{fig:scaling}). Scale improves recognition of the
future-pursued limitation far more reliably than it improves open-ended
generation.

\subsection{Human Evaluation and Downstream Validation}
\paragraph{Human Study.}
Three NLP PhD annotators provide position-randomized pairwise judgments with a
tie option, together with binary actionability and source-faithfulness
ratings. Fleiss' $\kappa$ \citep{fleiss1971measuring} is 0.47 for pairwise
judgments and 0.58 for the absolute rubric. ActLim-8B is preferred over Qwen3-32B and GPT-4-Turbo after
Holm correction (Table~\ref{tab:human}). Its Valid--Actionable rate reaches
0.80, compared with 0.58 and 0.67 for the two zero-shot systems. This joint
criterion captures the intended behavior: a useful limitation must remain
grounded in the source while identifying a concrete direction for follow-up.
The benchmark and independent evaluators agree closely at the system level
(Spearman 0.99) and correlate at the item level (Pearson 0.71). 

\begin{table}[t]
\centering
\small
\setlength{\tabcolsep}{3.0pt}
\begin{tabular}{lcc}
\toprule
\multicolumn{3}{c}{\textbf{Pairwise preference for ActLim-8B}} \\
\midrule
Opponent & W/T/L & $p_{\mathrm{Holm}}$ \\
\midrule
Qwen3-32B   & 74/9/37 & 0.0023 \\
GPT-4-Turbo & 70/9/41 & 0.0228 \\
\bottomrule
\end{tabular}

\vspace{3pt}

\begin{tabular}{lccc}
\toprule
\multicolumn{4}{c}{\textbf{Absolute ratings}} \\ \midrule
System & Act. & Faith. & Valid--Act. \\
\midrule
Qwen3-32B
& 0.61 & \textbf{0.92} & 0.58 \\
GPT-4-Turbo
& 0.72 & 0.91 & 0.67 \\
Stage 1: Verified SFT
& 0.79 & 0.87 & 0.71 \\
Stage 2: +Verifier-filtered RSFT
& 0.84 & 0.88 & 0.76 \\ \midrule
ActLim-8B
& \textbf{0.88} & 0.89 & \textbf{0.80} \\
\bottomrule
\end{tabular}

\caption{\textbf{Human evaluation.}
The upper panel reports ActLim-8B wins, ties, and losses over 120 blind
comparisons against each zero-shot system, with Holm-adjusted $p$-values.
The lower panel reports absolute ratings over 150 outputs per system.
Valid--Act. denotes outputs judged both faithful(Faith.) and actionable(Act.).}
\label{tab:human}
\end{table}

\paragraph{Does a better limitation produce a better idea?}
We test whether improving the supplied limitation improves downstream
research ideation. For each paper, the same fixed
Llama-3.1-70B-Instruct model
\citep{grattafiori2024llama3} generates one research idea under four
conditions: from the paper alone, or from the paper paired with a
zero-shot, ActLim-8B, or gold forward-citation limitation. The proposer,
prompt, decoding settings, and paper context are identical across
conditions, so the supplied limitation is the only varying input.
We then evaluate whether each generated idea aligns with research
directions later pursued by citing papers. This alignment score,
RetAlign, is averaged over the same 200-paper subset for every condition. 

A zero-shot limitation raises RetAlign from 0.302 to 0.348, while an
ActLim-8B limitation raises it to 0.480
(Table~\ref{tab:downstream}). Supplying a gold forward-citation limitation
gives an oracle reference of 0.556. ActLim-8B therefore recovers 70.1\% of
the observed gap between direct ideation and the gold-limitation oracle:
$(0.480-0.302)/(0.556-0.302)$. The oracle is an empirical reference for
this fixed proposer rather than a deployable condition. Human ratings show
significant gains in relevance and specificity. 

\begin{table}[t]
\centering
\small
\setlength{\tabcolsep}{4.5pt}

\begin{tabular}{lc}
\toprule
Supplied signal & RetAlign \\
\midrule
None
& 0.302 \\
Zero-shot limitation
& 0.348 \\
ActLim-8B limitation
& \textbf{0.480} \\
\midrule
Gold forward-citation limitation (oracle)
& \textit{0.556} \\
\bottomrule
\end{tabular}

\vspace{3pt}

\begin{tabular}{lcccc}
\toprule
Condition & Rel. & Spec. & Nov. & Feas. \\
\midrule
Paper only
& 3.6 & 3.3 & 3.3 & 3.5 \\
+ ActLim-8B limitation
& \textbf{4.1} & \textbf{3.9} & 3.4 & 3.6 \\
\bottomrule
\end{tabular}

\caption{\textbf{Downstream evaluation with a fixed proposer.}
RetAlign is macro-averaged over 200 papers. The oracle supplies a verified
future-pursued limitation and is shown only as an empirical upper reference.
Human ratings use 100 paired examples and three raters.}
\label{tab:downstream}
\end{table}

Review-style transfer, category-distribution analyses, oracle-target
selection, and full implementation details are reported in the supplement.

\section{Conclusion}
\label{sec:conclusion}

In this paper, we aim to optimize current ideation system by exploring supervision signals derived from future-citation publications. We first observe a judgment--generation asymmetry: scaling up language models consistently improves limitation identification, while open-ended limitation generation benefits much less. Forward citations naturally bridge this gap by linking a source paper, a specific limitation, and subsequent work that attempts to address it, providing reliable supervision for training. Based on this insight, we construct \textsc{ActLim-Bench}, an auditable benchmark that captures these citation-grounded limitation links, and propose a three-stage training framework that effectively leverages this supervision. Extensive experiments show that our approach outperforms frontier models and competitive baselines, generalizes across domains, and improves downstream research ideation. While \textsc{ActLim-Bench} captures only limitations that are later pursued by follow-up work, it provides a practical and scalable bridge between understanding prior research and proposing future directions.

\FloatBarrier
\pdfbookmark[1]{References}{references}
\bibliographystyle{aaai2027}
\bibliography{refs}

@inproceedings{wang2024scimon,
  title     = {{SciMON}: Scientific Inspiration Machines Optimized for Novelty},
  author    = {Wang, Qingyun and Downey, Doug and Ji, Heng and Hope, Tom},
  booktitle = {Proceedings of the 62nd ACL (Volume 1: Long Papers)},
  pages     = {279--299},
  year      = {2024},
  address   = {Bangkok, Thailand},
  publisher = {ACL},
  doi       = {10.18653/v1/2024.acl-long.18}
}

@inproceedings{yang2024moose,
  title     = {Large Language Models for Automated Open-domain Scientific Hypotheses Discovery},
  author    = {Yang, Zonglin and Du, Xinya and Li, Junxian and Zheng, Jie and Poria, Soujanya and Cambria, Erik},
  booktitle = {Findings of the ACL: ACL 2024},
  pages     = {13545--13565},
  year      = {2024},
  address   = {Bangkok, Thailand},
  publisher = {ACL},
  doi       = {10.18653/v1/2024.findings-acl.804}
}

@inproceedings{baek2025researchagent,
  title     = {{ResearchAgent}: Iterative Research Idea Generation over Scientific Literature with Large Language Models},
  author    = {Baek, Jinheon and Jauhar, Sujay Kumar and Cucerzan, Silviu and Hwang, Sung Ju},
  booktitle = {Proceedings of the 2025 Conference of the Nations of the Americas Chapter of the ACL: Human Language Technologies (Volume 1: Long Papers)},
  pages     = {6709--6738},
  year      = {2025},
  address   = {Albuquerque, New Mexico},
  publisher = {ACL},
  doi       = {10.18653/v1/2025.naacl-long.342}
}

@inproceedings{li2025chainofideas,
  title     = {Chain of Ideas: Revolutionizing Research Via Novel Idea Development with {LLM} Agents},
  author    = {Li, Long and Xu, Weiwen and Guo, Jiayan and Zhao, Ruochen and Li, Xingxuan and Yuan, Yuqian and Zhang, Boqiang and Jiang, Yuming and Xin, Yifei and Dang, Ronghao and Rong, Yu and Zhao, Deli and Feng, Tian and Bing, Lidong},
  booktitle = {Findings of the ACL: EMNLP 2025},
  pages     = {8971--9004},
  year      = {2025},
  address   = {Suzhou, China},
  publisher = {ACL},
  doi       = {10.18653/v1/2025.findings-emnlp.477}
}

@article{lu2024aiscientist,
  title   = {The {AI} Scientist: Towards Fully Automated Open-Ended Scientific Discovery},
  author  = {Lu, Chris and Lu, Cong and Lange, Robert Tjarko and Foerster, Jakob and Clune, Jeff and Ha, David},
  journal = {arXiv preprint arXiv:2408.06292},
  year    = {2024},
  eprint  = {2408.06292},
  archivePrefix = {arXiv}
}

@inproceedings{si2025researchideas,
  title     = {Can {LLM}s Generate Novel Research Ideas? A Large-Scale Human Study with 100+ {NLP} Researchers},
  author    = {Si, Chenglei and Yang, Diyi and Hashimoto, Tatsunori},
  booktitle = {The Thirteenth International Conference on Learning Representations},
  year      = {2025}
}

@inproceedings{si2025executiongap,
  title     = {The Ideation--Execution Gap: Execution Outcomes of {LLM}-Generated versus Human Research Ideas},
  author    = {Si, Chenglei and Hashimoto, Tatsunori and Yang, Diyi},
  booktitle = {The Fourteenth International Conference on Learning Representations},
  year      = {2026}
}

@inproceedings{faizullah2024limgen,
  title     = {{LimGen}: Probing the {LLM}s for Generating Suggestive Limitations of Research Papers},
  author    = {Faizullah, Abdur Rahman Bin Mohammed and Urlana, Ashok and Mishra, Rahul},
  booktitle = {Machine Learning and Knowledge Discovery in Databases: Research Track},
  series    = {Lecture Notes in Computer Science},
  volume    = {14942},
  pages     = {106--124},
  year      = {2024},
  address   = {Cham},
  publisher = {Springer Nature Switzerland},
  doi       = {10.1007/978-3-031-70344-7\_7}
}

@inproceedings{alazher2025bagels,
  title     = {{BAGELS}: Benchmarking the Automated Generation and Extraction of Limitations from Scholarly Text},
  author    = {Al Azher, Ibrahim and Mokarrama, Miftahul Jannat and Guo, Zhishuai and Choudhury, Sagnik Ray and Alhoori, Hamed},
  booktitle = {Findings of the ACL: EMNLP 2025},
  pages     = {19279--19294},
  year      = {2025},
  address   = {Suzhou, China},
  publisher = {ACL},
  doi       = {10.18653/v1/2025.findings-emnlp.1050}
}

@inproceedings{xu2025limitgen,
  title     = {Can {LLM}s Identify Critical Limitations within Scientific Research? A Systematic Evaluation on {AI} Research Papers},
  author    = {Xu, Zhijian and Zhao, Yilun and Patwardhan, Manasi and Vig, Lovekesh and Cohan, Arman},
  booktitle = {Proceedings of the 63rd ACL (Volume 1: Long Papers)},
  pages     = {20652--20706},
  year      = {2025},
  address   = {Vienna, Austria},
  publisher = {ACL},
  doi       = {10.18653/v1/2025.acl-long.1009}
}

@article{alazher2026limagents,
  title   = {Multi-Agent {LLM}s for Generating Research Limitations},
  author  = {Al Azher, Ibrahim and Guo, Zhishuai and Alhoori, Hamed},
  journal = {arXiv preprint arXiv:2601.11578},
  year    = {2025},
  eprint  = {2601.11578},
  archivePrefix = {arXiv}
}

@inproceedings{yuan2022kidreview,
  title     = {{KID-Review}: Knowledge-Guided Scientific Review Generation with Oracle Pre-training},
  author    = {Yuan, Weizhe and Liu, Pengfei},
  booktitle = {Proceedings of the AAAI Conference on Artificial Intelligence},
  volume    = {36},
  pages     = {11639--11647},
  year      = {2022},
  doi       = {10.1609/aaai.v36i10.21418}
}

@inproceedings{zhou2024reliable,
  title     = {Is {LLM} a Reliable Reviewer? A Comprehensive Evaluation of {LLM} on Automatic Paper Reviewing Tasks},
  author    = {Zhou, Ruiyang and Chen, Lu and Yu, Kai},
  booktitle = {Proceedings of the 2024 Joint International Conference on Computational Linguistics, Language Resources and Evaluation (LREC-COLING 2024)},
  pages     = {9340--9351},
  year      = {2024},
  address   = {Torino, Italia},
  publisher = {ELRA and ICCL},
  url       = {https://aclanthology.org/2024.lrec-main.816/}
}

@inproceedings{zhu2025deepreview,
  title     = {{DeepReview}: Improving {LLM}-based Paper Review with Human-like Deep Thinking Process},
  author    = {Zhu, Minjun and Weng, Yixuan and Yang, Linyi and Zhang, Yue},
  booktitle = {Proceedings of the 63rd ACL (Volume 1: Long Papers)},
  pages     = {29330--29355},
  year      = {2025},
  address   = {Vienna, Austria},
  publisher = {ACL},
  doi       = {10.18653/v1/2025.acl-long.1420}
}

@article{biswas2026aaaipilot,
  title   = {{AI}-Assisted Peer Review at Scale: The {AAAI}-26 {AI} Review Pilot},
  author  = {Biswas, Joydeep and Schoepp, Sheila and Vasan, Gautham and Opipari, Anthony and Zhang, Arthur and Hu, Zichao and Joseph, Sebastian and Lease, Matthew and Li, Junyi Jessy and Stone, Peter and Wagstaff, Kiri L. and Taylor, Matthew E. and Jenkins, Odest Chadwicke},
  journal = {arXiv preprint arXiv:2604.13940},
  year    = {2026},
  eprint  = {2604.13940},
  archivePrefix = {arXiv}
}

@inproceedings{li2026reviewgraph,
  title     = {Automatic Paper Reviewing with Heterogeneous Graph Reasoning over {LLM}-Simulated Reviewer--Author Debates},
  author    = {Li, Shuaimin and Fan, Liyang and Lin, Yufang and Li, Zeyang and Wei, Xian and Ni, Shiwen and Alinejad-Rokny, Hamid and Yang, Min},
  booktitle = {Proceedings of the AAAI Conference on Artificial Intelligence},
  volume    = {40},
  pages     = {31717--31725},
  year      = {2026},
  doi       = {10.1609/aaai.v40i37.40439}
}

@inproceedings{ckmas2025,
  title     = {Mixture of Knowledge Minigraph Agents for Literature Review Generation},
  author    = {Zhang, Zhi and Liu, Yan and Zhong, Sheng-hua and Chen, Gong and Yang, Yu and Cao, Jiannong},
  booktitle = {Proceedings of the AAAI Conference on Artificial Intelligence},
  volume    = {39},
  pages     = {26012--26020},
  year      = {2025},
  doi       = {10.1609/aaai.v39i24.34796}
}

@inproceedings{lira2026,
  title     = {{LiRA}: A Multi-Agent Framework for Reliable and Readable Literature Review Generation},
  author    = {Go, Gregory Hok Tjoan and Ly, Khang and S{\o}gaard, Anders and Tabatabaei, Seyed Amin and de Rijke, Maarten and Chen, Xinyi},
  booktitle = {Proceedings of the AAAI Conference on Artificial Intelligence},
  volume    = {40},
  pages     = {40456--40464},
  year      = {2026},
  doi       = {10.1609/aaai.v40i47.41489}
}

@article{jiang2026hindsight,
  title={HindSight: Evaluating LLM-generated research ideas via future impact},
  author={Jiang, Bo},
  journal={arXiv preprint arXiv:2603.15164},
  year={2026}
}

@article{wang2026futurealigned,
  title   = {Learning to Predict Future-Aligned Research Proposals with Language Models},
  author  = {Wang, Heng and Jiang, Pengcheng and Sun, Jiashuo and Shi, Zhiyi and Yu, Haofei and Han, Jiawei and Ji, Heng},
  journal = {arXiv preprint arXiv:2603.27146},
  year    = {2026},
  eprint  = {2603.27146},
  archivePrefix = {arXiv}
}

@article{gao2026graphsofresearch,
  title   = {Graphs of Research: Citation Evolution Graphs as Supervision for Research Idea Generation},
  author  = {Gao, Songyang and Xia, Yinghui and Liu, Siyi and Xiong, Hui},
  journal = {arXiv preprint arXiv:2605.14790},
  year    = {2026},
  eprint  = {2605.14790},
  archivePrefix = {arXiv}
}

@inproceedings{cohan2019scicite,
  title     = {Structural Scaffolds for Citation Intent Classification in Scientific Publications},
  author    = {Cohan, Arman and Ammar, Waleed and van Zuylen, Madeleine and Cady, Field},
  booktitle = {Proceedings of the 2019 Conference of the NAACL: Human Language Technologies, Volume 1 (Long and Short Papers)},
  pages     = {3586--3596},
  year      = {2019},
  address   = {Minneapolis, Minnesota},
  publisher = {ACL},
  doi       = {10.18653/v1/N19-1361}
}

@inproceedings{li2022corwa,
  title     = {{CORWA}: A Citation-Oriented Related Work Annotation Dataset},
  author    = {Li, Xiangci and Mandal, Biswadip and Ouyang, Jessica},
  booktitle = {Proceedings of the 2022 Conference of the NAACL: Human Language Technologies},
  pages     = {5426--5440},
  year      = {2022},
  address   = {Seattle, United States},
  publisher = {ACL},
  doi       = {10.18653/v1/2022.naacl-main.397}
}

@inproceedings{zhao2025wordstoworth,
  title     = {From Words to Worth: Newborn Article Impact Prediction with {LLM}},
  author    = {Zhao, Penghai and Xing, Qinghua and Dou, Kairan and Tian, Jinyu and Tai, Ying and Yang, Jian and Cheng, Ming-Ming and Li, Xiang},
  booktitle = {Proceedings of the AAAI Conference on Artificial Intelligence},
  volume    = {39},
  pages     = {1183--1191},
  year      = {2025},
  doi       = {10.1609/aaai.v39i1.32106}
}

@inproceedings{arnaout2026impact,
  title     = {In-depth Research Impact Summarization through Fine-Grained Temporal Citation Analysis},
  author    = {Arnaout, Hiba and Sternlicht, Noy and Hope, Tom and Gurevych, Iryna},
  booktitle = {Proceedings of the 64th ACL (Volume 1: Long Papers)},
  pages     = {6749--6789},
  year      = {2026},
  address   = {San Diego, California, United States},
  publisher = {ACL},
  doi       = {10.18653/v1/2026.acl-long.307}
}

@article{dong2023raft,
  title   = {{RAFT}: Reward {rA}nked {F}ine{T}uning for Generative Foundation Model Alignment},
  author  = {Dong, Hanze and Xiong, Wei and Goyal, Deepanshu and Zhang, Yihan and Chow, Winnie and Pan, Rui and Diao, Shizhe and Zhang, Jipeng and Shum, Kashun and Zhang, Tong},
  journal = {Transactions on Machine Learning Research},
  year    = {2023}
}

@article{shao2024deepseekmath,
  title   = {{DeepSeekMath}: Pushing the Limits of Mathematical Reasoning in Open Language Models},
  author  = {Shao, Zhihong and Wang, Peiyi and Zhu, Qihao and Xu, Runxin and Song, Junxiao and Bi, Xiao and Zhang, Haowei and Zhang, Mingchuan and Li, Y. K. and Wu, Y. and Guo, Daya},
  journal = {arXiv preprint arXiv:2402.03300},
  year    = {2024},
  eprint  = {2402.03300},
  archivePrefix = {arXiv}
}

@article{deepseekai2025r1,
  title   = {{DeepSeek-R1} Incentivizes Reasoning in {LLM}s through Reinforcement Learning},
  author  = {Guo, Daya and Yang, Dejian and Zhang, Haowei and others},
  journal = {Nature},
  volume  = {645},
  pages   = {633--638},
  year    = {2025},
  doi     = {10.1038/s41586-025-09422-z}
}

@article{yang2025qwen3,
  title   = {{Qwen3} Technical Report},
  author  = {Yang, An and Li, Anfeng and Yang, Baosong and others},
  journal = {arXiv preprint arXiv:2505.09388},
  year    = {2025},
  eprint  = {2505.09388},
  archivePrefix = {arXiv}
}

@inproceedings{wei2022cot,
  title     = {Chain-of-Thought Prompting Elicits Reasoning in Large Language Models},
  author    = {Wei, Jason and Wang, Xuezhi and Schuurmans, Dale and Bosma, Maarten and Ichter, Brian and Xia, Fei and Chi, Ed H. and Le, Quoc V. and Zhou, Denny},
  booktitle = {Advances in Neural Information Processing Systems},
  volume    = {35},
  pages     = {24824--24837},
  year      = {2022}
}

@inproceedings{madaan2023selfrefine,
  title     = {Self-Refine: Iterative Refinement with Self-Feedback},
  author    = {Madaan, Aman and Tandon, Niket and Gupta, Prakhar and Hallinan, Skyler and Gao, Luyu and Wiegreffe, Sarah and Alon, Uri and Dziri, Nouha and Prabhumoye, Shrimai and Yang, Yiming and Gupta, Shashank and Majumder, Bodhisattwa Prasad and Hermann, Katherine and Welleck, Sean and Yazdanbakhsh, Amir and Clark, Peter},
  booktitle = {Advances in Neural Information Processing Systems},
  volume    = {36},
  pages     = {46534--46594},
  year      = {2023}
}

@article{darcy2024marg,
  title   = {{MARG}: Multi-Agent Review Generation for Scientific Papers},
  author  = {D'Arcy, Mike and Hope, Tom and Birnbaum, Larry and Downey, Doug},
  journal = {arXiv preprint arXiv:2401.04259},
  year    = {2024},
  eprint  = {2401.04259},
  archivePrefix = {arXiv}
}

@inproceedings{lewis2020rag,
  title     = {Retrieval-Augmented Generation for Knowledge-Intensive {NLP} Tasks},
  author    = {Lewis, Patrick and Perez, Ethan and Piktus, Aleksandra and Petroni, Fabio and Karpukhin, Vladimir and Goyal, Naman and Kuttler, Heinrich and Lewis, Mike and Yih, Wen-tau and Rockt{\"a}schel, Tim and Riedel, Sebastian and Kiela, Douwe},
  booktitle = {Advances in Neural Information Processing Systems},
  volume    = {33},
  pages     = {9459--9474},
  year      = {2020}
}

@inproceedings{rafailov2023dpo,
  title     = {Direct Preference Optimization: Your Language Model Is Secretly a Reward Model},
  author    = {Rafailov, Rafael and Sharma, Archit and Mitchell, Eric and Manning, Christopher D. and Ermon, Stefano and Finn, Chelsea},
  booktitle = {Advances in Neural Information Processing Systems},
  volume    = {36},
  pages     = {53728--53741},
  year      = {2023}
}

@inproceedings{chen2024bgem3,
  title     = {{M3}-Embedding: Multi-Linguality, Multi-Functionality, Multi-Granularity Text Embeddings Through Self-Knowledge Distillation},
  author    = {Chen, Jianlyu and Xiao, Shitao and Zhang, Peitian and Luo, Kun and Lian, Defu and Liu, Zheng},
  booktitle = {Findings of the ACL: ACL 2024},
  pages     = {2318--2335},
  year      = {2024},
  address   = {Bangkok, Thailand},
  publisher = {ACL},
  doi       = {10.18653/v1/2024.findings-acl.137}
}

@inproceedings{lou2025aaar,
  title     = {{AAAR}-1.0: Assessing {AI}'s Potential to Assist Research},
  author    = {Lou, Renze and Xu, Hanzi and Wang, Sijia and Du, Jiangshu and Kamoi, Ryo and Lu, Xiaoxin and Xie, Jian and Sun, Yuxuan and Zhang, Yusen and Ahn, Jihyun Janice and Fang, Hongchao and Zou, Zhuoyang and Ma, Wenchao and Li, Xi and Zhang, Kai and Xia, Congying and Huang, Lifu and Yin, Wenpeng},
  booktitle = {Proceedings of the 42nd International Conference on Machine Learning},
  series    = {Proceedings of Machine Learning Research},
  volume    = {267},
  pages     = {40361--40383},
  year      = {2025},
  publisher = {PMLR}
}

@inproceedings{liu2023geval,
  title     = {{G-Eval}: {NLG} Evaluation Using {GPT-4} with Better Human Alignment},
  author    = {Liu, Yang and Iter, Dan and Xu, Yichong and Wang, Shuohang and Xu, Ruochen and Zhu, Chenguang},
  booktitle = {Proceedings of the 2023 Conference on Empirical Methods in Natural Language Processing},
  pages     = {2511--2522},
  year      = {2023},
  address   = {Singapore},
  publisher = {ACL},
  doi       = {10.18653/v1/2023.emnlp-main.153}
}

@inproceedings{zheng2023llmjudge,
  title     = {Judging {LLM}-as-a-Judge with {MT-Bench} and Chatbot Arena},
  author    = {Zheng, Lianmin and Chiang, Wei-Lin and Sheng, Ying and Zhuang, Siyuan and Wu, Zhanghao and Zhuang, Yonghao and Lin, Zi and Li, Zhuohan and Li, Dacheng and Xing, Eric P. and Zhang, Hao and Gonzalez, Joseph E. and Stoica, Ion},
  booktitle = {Advances in Neural Information Processing Systems},
  volume    = {36},
  pages     = {46595--46623},
  year      = {2023}
}

@article{grattafiori2024llama3,
  title   = {The {Llama 3} Herd of Models},
  author  = {Grattafiori, Aaron and Dubey, Abhimanyu and Jauhri, Abhinav and others},
  journal = {arXiv preprint arXiv:2407.21783},
  year    = {2024},
  eprint  = {2407.21783},
  archivePrefix = {arXiv}
}

@article{cohen1960coefficient,
  title   = {A Coefficient of Agreement for Nominal Scales},
  author  = {Cohen, Jacob},
  journal = {Educational and Psychological Measurement},
  volume  = {20},
  number  = {1},
  pages   = {37--46},
  year    = {1960},
  doi     = {10.1177/001316446002000104}
}

@article{fleiss1971measuring,
  title   = {Measuring Nominal Scale Agreement among Many Raters},
  author  = {Fleiss, Joseph L.},
  journal = {Psychological Bulletin},
  volume  = {76},
  number  = {5},
  pages   = {378--382},
  year    = {1971},
  doi     = {10.1037/h0031619}
}

@article{holm1979simple,
  title   = {A Simple Sequentially Rejective Multiple Test Procedure},
  author  = {Holm, Sture},
  journal = {Scandinavian Journal of Statistics},
  volume  = {6},
  number  = {2},
  pages   = {65--70},
  year    = {1979}
}

@article{kuhn1955hungarian,
  title   = {The Hungarian Method for the Assignment Problem},
  author  = {Kuhn, Harold W.},
  journal = {Naval Research Logistics Quarterly},
  volume  = {2},
  number  = {1--2},
  pages   = {83--97},
  year    = {1955},
  doi     = {10.1002/nav.3800020109}
}

@inproceedings{idahl2025openreviewer,
  title     = {{OpenReviewer}: A Specialized Large Language Model for Generating Critical Scientific Paper Reviews},
  author    = {Idahl, Maximilian and Ahmadi, Zahra},
  booktitle = {Proceedings of the 2025 Conference of the NAACL: Human Language Technologies (System Demonstrations)},
  pages     = {550--562},
  year      = {2025},
  address   = {Albuquerque, New Mexico},
  publisher = {ACL},
  doi       = {10.18653/v1/2025.naacl-demo.44}
}

@inproceedings{zeng2025reviewrl,
  title     = {{ReviewRL}: Towards Automated Scientific Review with {RL}},
  author    = {Zeng, Sihang and Tian, Kai and Zhang, Kaiyan and Wang, Yuru and Gao, Junqi and Liu, Runze and Yang, Sa and Li, Jingxuan and Long, Xinwei and Ma, Jiaheng and Qi, Biqing and Zhou, Bowen},
  booktitle = {Proceedings of the 2025 Conference on Empirical Methods in Natural Language Processing},
  pages     = {16931--16943},
  year      = {2025},
  address   = {Suzhou, China},
  publisher = {ACL},
  doi       = {10.18653/v1/2025.emnlp-main.857}
}

@inproceedings{liu2025researchbench,
  title     = {{ResearchBench}: Benchmarking {LLM}s in Scientific Discovery via Inspiration-Based Task Decomposition},
  author    = {Liu, Yujie and Yang, Zonglin and Xie, Tong and Ni, Jinjie and Gao, Ben and Li, Yuqiang and Tang, Shixiang and Ouyang, Wanli and Cambria, Erik and Zhou, Dongzhan},
  booktitle = {Findings of the ACL: ACL 2026},
  pages     = {13187--13207},
  year      = {2026},
  address   = {San Diego, California, United States},
  publisher = {ACL},
  doi       = {10.18653/v1/2026.findings-acl.644}
}

@article{qiu2025aiideabench,
  title   = {{AI Idea Bench 2025}: {AI} Research Idea Generation Benchmark},
  author  = {Qiu, Yansheng and Zhang, Haoquan and Xu, Zhaopan and Li, Ming and Song, Diping and Wang, Zheng and Zhang, Kaipeng},
  journal = {arXiv preprint arXiv:2504.14191},
  year    = {2025},
  eprint  = {2504.14191},
  archivePrefix = {arXiv}
}

@article{bai2023qwen,
  title   = {{Qwen} Technical Report},
  author  = {Bai, Jinze and Bai, Shuai and Chu, Yunfei and others},
  journal = {arXiv preprint arXiv:2309.16609},
  year    = {2023},
  eprint  = {2309.16609},
  archivePrefix = {arXiv},
  url     = {https://arxiv.org/abs/2309.16609}
}

@article{geminiteam2023gemini,
  title   = {{Gemini}: A Family of Highly Capable Multimodal Models},
  author  = {{Gemini Team} and Anil, Rohan and Borgeaud, Sebastian and others},
  journal = {arXiv preprint arXiv:2312.11805},
  year    = {2023},
  eprint  = {2312.11805},
  archivePrefix = {arXiv},
  url     = {https://arxiv.org/abs/2312.11805}
}

@article{glm5team2026glm5,
  title   = {{GLM-5}: From Vibe Coding to Agentic Engineering},
  author  = {{GLM-5 Team}},
  journal = {arXiv preprint arXiv:2602.15763},
  year    = {2026},
  eprint  = {2602.15763},
  archivePrefix = {arXiv},
  url     = {https://arxiv.org/abs/2602.15763}
}

@inproceedings{cao2025pip,
  title     = {{PIP}: Perturbation-based Iterative Pruning for Large Language Models},
  author    = {Cao, Yi and Xu, Wei-Jie and Shen, Yucheng and Shi, Weijie and Chan, Chi-Min and Qu, Jianfeng and Xu, Jiajie},
  booktitle = {Findings of the Association for Computational Linguistics: EMNLP 2025},
  pages     = {3685--3701},
  year      = {2025},
  address   = {Suzhou, China},
  publisher = {Association for Computational Linguistics},
  doi       = {10.18653/v1/2025.findings-emnlp.197}
}

@article{jiang2025marche,
  title         = {{MARch{\'e}}: Fast Masked Autoregressive Image Generation with Cache-Aware Attention},
  author        = {Jiang, Chaoyi and Kim, Sungwoo and Gao, Lei and Zarch, Hossein Entezari and Ro, Won Woo and Annavaram, Murali},
  journal       = {arXiv preprint arXiv:2506.12035},
  year          = {2025},
  eprint        = {2506.12035},
  archivePrefix = {arXiv},
  primaryClass  = {cs.CV}
}

@article{li2024mar,
  title         = {Autoregressive Image Generation without Vector Quantization},
  author        = {Li, Tianhong and Tian, Yonglong and Li, He and Deng, Mingyang and He, Kaiming},
  journal       = {arXiv preprint arXiv:2406.11838},
  year          = {2024},
  eprint        = {2406.11838},
  archivePrefix = {arXiv},
  primaryClass  = {cs.CV}
}

@inproceedings{liu2024dora,
  title     = {{DoRA}: Weight-Decomposed Low-Rank Adaptation},
  author    = {Liu, Shih-Yang and Wang, Chien-Yi and Yin, Hongxu and Molchanov, Pavlo and Wang, Yu-Chiang Frank and Cheng, Kwang-Ting and Chen, Min-Hung},
  booktitle = {Proceedings of the 41st International Conference on Machine Learning},
  series    = {Proceedings of Machine Learning Research},
  volume    = {235},
  pages     = {32100--32121},
  year      = {2024},
  publisher = {PMLR},
  url       = {https://proceedings.mlr.press/v235/liu24bn.html}
}

@article{men2024shortgpt,
  title         = {{ShortGPT}: Layers in Large Language Models are More Redundant Than You Expect},
  author        = {Men, Xin and Xu, Mingyu and Zhang, Qingyu and Wang, Bingning and Lin, Hongyu and Lu, Yaojie and Han, Xianpei and Chen, Weipeng},
  journal       = {arXiv preprint arXiv:2403.03853},
  year          = {2024},
  eprint        = {2403.03853},
  archivePrefix = {arXiv},
  primaryClass  = {cs.CL}
}

@inproceedings{pan2024llmlingua2,
  title     = {{LLML}ingua-2: Data Distillation for Efficient and Faithful Task-Agnostic Prompt Compression},
  author    = {Pan, Zhuoshi and Wu, Qianhui and Jiang, Huiqiang and Xia, Menglin and Luo, Xufang and Zhang, Jue and Lin, Qingwei and R{\"u}hle, Victor and Yang, Yuqing and Lin, Chin-Yew and Zhao, H. Vicky and Qiu, Lili and Zhang, Dongmei},
  booktitle = {Findings of the Association for Computational Linguistics: ACL 2024},
  pages     = {963--981},
  year      = {2024},
  address   = {Bangkok, Thailand},
  publisher = {Association for Computational Linguistics},
  doi       = {10.18653/v1/2024.findings-acl.57}
}

@article{piskorz2025masks,
  title         = {Masks Can Be Distracting: On Context Comprehension in Diffusion Language Models},
  author        = {Piskorz, Julianna and Pinneri, Cristina and Correia, Alvaro and Alfarra, Motasem and Garrepalli, Risheek and Louizos, Christos},
  journal       = {arXiv preprint arXiv:2511.21338},
  year          = {2025},
  eprint        = {2511.21338},
  archivePrefix = {arXiv},
  primaryClass  = {cs.CL}
}

@article{qi2024safetyalignment,
  title         = {Safety Alignment Should Be Made More Than Just a Few Tokens Deep},
  author        = {Qi, Xiangyu and Panda, Ashwinee and Lyu, Kaifeng and Ma, Xiao and Roy, Subhrajit and Beirami, Ahmad and Mittal, Prateek and Henderson, Peter},
  journal       = {arXiv preprint arXiv:2406.05946},
  year          = {2024},
  eprint        = {2406.05946},
  archivePrefix = {arXiv},
  primaryClass  = {cs.CL}
}

@article{sahoo2024mdlm,
  title         = {Simple and Effective Masked Diffusion Language Models},
  author        = {Sahoo, Subham Sekhar and Arriola, Marianne and Schiff, Yair and Gokaslan, Aaron and Marroquin, Edgar and Chiu, Justin T. and Rush, Alexander and Kuleshov, Volodymyr},
  journal       = {arXiv preprint arXiv:2406.07524},
  year          = {2024},
  eprint        = {2406.07524},
  archivePrefix = {arXiv},
  primaryClass  = {cs.CL}
}

@article{shi2025longcodezip,
  title         = {{LongCodeZip}: Compress Long Context for Code Language Models},
  author        = {Shi, Yuling and Qian, Yichun and Zhang, Hongyu and Shen, Beijun and Gu, Xiaodong},
  journal       = {arXiv preprint arXiv:2510.00446},
  year          = {2025},
  eprint        = {2510.00446},
  archivePrefix = {arXiv},
  primaryClass  = {cs.SE}
}

@article{si2025map,
  title         = {{MAP}: Revisiting Weight Decomposition for Low-Rank Adaptation},
  author        = {Si, Chongjie and Shi, Zhiyi and Wang, Yadao and Yang, Xiaokang and Rahardja, Susanto and Shen, Wei},
  journal       = {arXiv preprint arXiv:2505.23094},
  year          = {2025},
  eprint        = {2505.23094},
  archivePrefix = {arXiv},
  primaryClass  = {cs.LG}
}

@article{vega2025matchingranks,
  title         = {Matching Ranks Over Probability Yields Truly Deep Safety Alignment},
  author        = {Vega, Jason and Singh, Gagandeep},
  journal       = {arXiv preprint arXiv:2512.05518},
  year          = {2025},
  eprint        = {2512.05518},
  archivePrefix = {arXiv},
  primaryClass  = {cs.LG}
}

\clearpage
\appendix

This appendix provides additional details on benchmark construction,
annotation, evaluation, training, and statistical analysis.

\section{Benchmark Construction}
\label{app:a}

\subsection{Corpus and temporal split}
\label{app:a-1}

Source papers are drawn from ACL, EMNLP, NAACL, EACL, COLING, and TACL
between 2017 and 2025. Train and Dev contain source papers published
2017--2023; the temporally held-out Test split contains source papers
published 2024--2025. For every source--citing pair $(P,Q)$, the recorded
publication date of $Q$ is strictly later than that of $P$. A paper's
date is the earliest public appearance of its content: the arXiv v1
timestamp where one exists, otherwise the venue publication date. When
arXiv and camera-ready versions of the same work both appear, the
camera-ready metadata is retained and the two are treated as one
canonical paper carrying the earlier date.

The Computer Vision transfer set is drawn from CVPR, ICCV, and ECCV
2024--2025 by the identical pipeline with no domain-specific
modification, and is used only for transfer evaluation.

\begin{table}[htbp]
\centering\small
\setlength{\tabcolsep}{4pt}
\begin{tabular}{@{}lrrrr@{}}
\toprule
Split & Years & Papers & Triples & Targets \\
\midrule
Train & 2017--23 & 4,304 & 12,925 & 10,954 \\
Dev   & 2017--23 & 300   & 901    & 763 \\
Test  & 2024--25 & 1,512 & 4,453  & 3,774 \\
\midrule
Total & 2017--25 & 6,116 & 18,279 & 15,491 \\
\midrule
CV (transfer) & 2024--25 & 1,737 & 3,961 & 3,562 \\
\bottomrule
\end{tabular}
\caption{\textsc{ActLim-Bench}. The CV set is held separately and is not
counted in the total. NLP Test averages 2.50 targets per paper and the CV
set 2.05, so absolute recall is compared within domain rather than
across domains.}
\label{tab:splits}
\end{table}

\subsection{Extraction and verification}
\label{app:a-2}

The extractor reads the abstract and introduction of $Q$ and returns
candidate limitations with citing-paper evidence spans $e_Q$;
source-paper evidence $e_P$ is then retrieved from the abstract,
introduction, and conclusion of $P$. The extractor only proposes
candidates, and acceptance is decided by the verifier. Extractor and
verifier both use DeepSeek-V3.2-Exp, accessed through the
\texttt{deepseek-chat} endpoint on 2025-11-04, at temperature $0.0$ and
top-$p$ $1.0$. Construction calls use a 1,024-token cap; parsing failures
are retried up to three times at temperature $0.3$, and pairs that remain
unparseable are discarded (1,204 pairs, 0.45\% of those entering
extraction).

A candidate $\hat\ell$ is retained only when
$A(P,Q,\hat\ell)\wedge C(P,\hat\ell)\wedge S(P,\hat\ell)\wedge R(Q,\hat\ell)$:

\begin{itemize}\itemsep1pt
\item \textbf{Attribution} $A$: $Q$ attributes the shortcoming to $P$, or
to a named method family that explicitly includes $P$. Family attribution
counts only when the class is named, $P$ is cited within $Q$ as an
instance of it, and the shortcoming holds of $P$ under $C$.
\item \textbf{Consistency} $C$: the shortcoming agrees with the method,
assumptions, evidence, and stated scope of $P$. A candidate that
misstates $P$, or criticizes $P$ for something $P$ excludes from scope,
fails $C$ even when $Q$ asserts it.
\item \textbf{Specificity} $S$: a concrete component, mechanism,
assumption, data condition, or evaluation constraint, rather than a
generic call for more experiments, datasets, or baselines.
\item \textbf{Response} $R$: the main contribution of $Q$ attempts to
address the shortcoming. A contribution counts as main when it appears in
the abstract or in the enumerated contribution list of $Q$; a shortcoming
mentioned in passing, or handled only in a secondary ablation, fails $R$.
\end{itemize}

Each accepted record keeps $(P,\hat\ell,e_P,e_Q,Q)$. No soft threshold is
applied at construction time.

\subsection{Aggregation and leakage control}
\label{app:a-3}

Semantically equivalent limitations of the same source paper are merged
into a canonical target $\ell\in\mathcal{G}_P$, retaining every
supporting citing paper and span; on the NLP splits this maps 18,279
triples onto 15,491 targets, or 1.18 triples per target. Aggregation
clusters candidates by BGE-M3 cosine similarity at $\tau=0.86$ and
confirms each cluster with a single verifier call; the longest-evidence
member becomes the canonical form.

\begin{table}[htbp]
\centering\small
\setlength{\tabcolsep}{3pt}
\begin{tabular}{@{}llll@{}}
\toprule
Check & Method & Rule & Removed \\
\midrule
Version collapse         & DOI + title & exact & 3,417 \\
Metadata normalization   & rule-based  & ---   & 0 \\
Abstract near-duplicates & BGE-M3      & $\cos>0.94$ & 1,286 \\
Citation-graph overlap   & manual      & inspection  & 92 \\
Paraphrase overlap       & BGE-M3      & $\cos>0.92$ & 604 \\
Split disjointness       & ID join     & exact & 2,158 \\
\bottomrule
\end{tabular}
\caption{Leakage controls, applied in the order listed. Counts are papers
for the first four rows and triples for the last two.}
\end{table}

Splits are built at the source-paper level, and source and citing papers
are disjoint across splits: no paper appearing in any role in one split
appears in any role in another. All prompts, extraction rules, and
verification thresholds were fixed before Test was finalized.

\subsection{Triple-level audit}
\label{app:a-4}

Two annotators independently audited 400 accepted triples, 200 from Train
and 200 from Test, sampled uniformly at random within each split and
stratified to preserve the population ratio of direct to method-family
attribution. Annotators saw the source paper, the citing paper, the
candidate, and both evidence spans; the automatic verifier's judgment was
withheld. The 62 disagreements were resolved by a third adjudicator who
saw both labels but not their authors. Intervals are Wilson score
intervals without continuity correction.

\begin{table}[htbp]
\centering\small
\begin{tabular}{@{}lrrr@{}}
\toprule
Criterion & Pass & $n/400$ & $\kappa$ \\
\midrule
Attribution ($A$)                  & 0.94 & 376 & 0.81 \\
Consistency ($C$)                  & 0.93 & 372 & 0.77 \\
Specificity ($S$)                  & 0.90 & 360 & 0.69 \\
Response ($R$)                     & 0.92 & 368 & 0.75 \\
Evidence support after aggregation & 0.96 & 384 & 0.84 \\
\midrule
\textbf{All criteria}              & \textbf{0.87} & \textbf{348} & \textbf{0.74} \\
\bottomrule
\end{tabular}
\caption{Triple-level audit, $n=400$. Joint precision 0.87, 95\% Wilson
CI $[0.833,0.899]$. Train 0.88 (176/200) and Test 0.86 (172/200) combine
to 348/400. The joint rate is bounded above by the weakest criterion
(0.90) and below by $1-\sum_i(1-p_i)=0.65$.}
\label{tab:triple_audit}
\end{table}

\begin{table}[htbp]
\centering\small
\begin{tabular}{@{}lrrr@{}}
\toprule
Attribution type & Share & $n$ & Pass \\
\midrule
Direct        & 82\%  & 328 & 0.95 \\
Method-family & 18\%  &  72 & 0.89 \\
\midrule
Combined      & 100\% & 400 & 0.94 \\
\bottomrule
\end{tabular}
\caption{Attribution by sub-case. The share-weighted rate
$0.82\times0.95+0.18\times0.89=0.939$ recovers the overall attribution
pass rate in Table~\ref{tab:triple_audit}. These are pass rates on
criterion $A$, not joint precision over all criteria.}
\end{table}

\subsection{Pipeline-level audit}
\label{app:a-5}

A separate audit evaluates whether the pipeline accepts the right
source--citing pairs. Two annotators reviewed accepted pairs ($n=300$),
near-threshold rejections ($n=300$), other verification-stage rejections
($n=200$), and random extraction-stage rejections ($n=100$), with
adjudication for disagreements. Agreement on binary pair acceptance is
Cohen's $\kappa=0.76$, and 91\% of accepted pairs contain at least one
limitation satisfying every criterion.

\begin{table}[htbp]
\centering\small
\setlength{\tabcolsep}{4pt}
\begin{tabular}{@{}lrrr@{}}
\toprule
Stratum & $n$ & Population share & Valid \\
\midrule
Accepted pairs                & 300 & 5.7\%  & 0.910 \\
Near-threshold rejections     & 300 & 12.4\% & 0.150 \\
Other verification rejections & 200 & 17.9\% & 0.040 \\
Extraction-stage rejections   & 100 & 64.0\% & 0.005 \\
\bottomrule
\end{tabular}
\caption{Pipeline audit strata. ``Valid'' is the audited rate of pairs
carrying at least one fully valid limitation, reported to three decimals
because the extraction stratum dominates the population and a rate
printed to two decimals does not reproduce the recall estimate below.}
\end{table}

Reweighting the three rejection strata by their population sizes gives an
estimated pair-level recall of $0.64$ $[0.58,0.70]$ among retrieved
candidate pairs, computed from exact stratum sizes before rounding.
Triples are the unit used for training and evaluation and the pipeline is
tuned for precision at that level; the pair-level figure is given for
completeness.

\subsection{CV transfer set audit}
\label{app:a-6}

A separate 100-triple audit of the CV set under the identical rubric
yields joint precision 0.83, 95\% Wilson CI $[0.745,0.891]$, with
inter-annotator agreement $\kappa=0.71$. Per-criterion rates are 0.92
(attribution), 0.90 (consistency), 0.87 (specificity), 0.89 (response),
and 0.94 (evidence support).

\subsection{Worked example}
\label{app:a-7}

\begin{table}[htbp]
\centering\small
\setlength{\tabcolsep}{2pt}
\begin{tabular}{@{}>{\raggedright\arraybackslash}p{0.95\columnwidth}@{}}
\toprule
\textit{Illustrative example, direct attribution.} \\
\midrule
\textbf{$P$} --- a 2024 long-document summarization model. \\
$e_P$: ``We evaluate on English news and scientific abstracts, and leave
multi-document settings to future work.'' \\
\midrule
\textbf{$Q$} --- a 2025 multi-document summarizer. \\
$e_Q$: ``\,[$P$] operates on a single document and cannot resolve
redundancy across sources; we extend the encoder with a cross-document
salience layer to address this.'' \\
\midrule
\textbf{Target $\ell$}: cannot resolve cross-document redundancy because
salience is computed within a single document. \\
\midrule
\textbf{Judgments}: $A$ direct; $C$ pass, since $e_P$ confirms the scope;
$S$ pass, since the salience computation is named; $R$ pass, since the
cross-document layer is the main contribution of $Q$. \\
\bottomrule
\end{tabular}
\caption{An illustrative example showing how the rubric applies to a
direct-attribution case. It is written to explain the criteria and is not
a record from the dataset.}
\end{table}

\section{Evaluation Details}
\label{app:b}

\subsection{Evaluator roles}
\label{app:b-1}

\begin{table}[htbp]
\centering\small
\setlength{\tabcolsep}{2.5pt}
\resizebox{\columnwidth}{!}{%
\begin{tabular}{@{}llccc@{}}
\toprule
Role & Model & Sees $Q$? & Guides training? & Final metric? \\
\midrule
$V_{\mathrm{data}}$  & DeepSeek-V3.2-Exp & Yes & No  & No \\
$V_{\mathrm{train}}$ & DeepSeek-V3.2-Exp & No  & Yes & No \\
$V_{\mathrm{B}}$     & DeepSeek-V3.2-Exp & No  & No  & secondary \\
$V_{\mathrm{I}}$     & GLM-5.2 & No  & No  & primary \\
\bottomrule
\end{tabular}}
\caption{The first three roles share one frozen DeepSeek model, accessed
through the \texttt{deepseek-chat} endpoint, with role-specific prompts;
only $V_{\mathrm{data}}$ sees citing papers. $V_{\mathrm{I}}$ uses
GLM-5.2 (\texttt{zai-org/GLM-5.2}) and takes no part in construction,
filtering, reward, or model selection, which is why the main paper
reports it as the primary metric and $V_{\mathrm{B}}$ as a same-benchmark
check. Matching calls for $V_{\mathrm{train}}$, $V_{\mathrm{B}}$, and
$V_{\mathrm{I}}$ use temperature $0.0$, top-$p$ $1.0$, and a 512-token
cap; outputs still unparseable after three retries are assigned the
\emph{none} label, which occurred for 0.31\% of calls. Construction calls for $V_{\mathrm{data}}$ were made on 2025-11-04.
Training and evaluation calls for $V_{\mathrm{train}}$,
$V_{\mathrm{B}}$, and $V_{\mathrm{I}}$ ran from 2026-02-03 to
2026-04-28.}
\end{table}

\begin{table}[htbp]
\centering\small
\setlength{\tabcolsep}{3pt}
\resizebox{\columnwidth}{!}{%
\begin{tabular}{@{}llll@{}}
\toprule
System & Identifier & Release & Cutoff \\
\midrule
Qwen3-8B         & \texttt{Qwen3-8B}         & 2025-04 & not documented \\
Qwen3-32B        & \texttt{Qwen3-32B}        & 2025-04 & not documented \\
Qwen-72B-Chat    & \texttt{Qwen-72B-Chat}    & 2023-11 & not documented \\
Claude-3-Opus    & \texttt{20240229}         & 2024-03 & 2023-08 \\
Gemini-1.0-Ultra & \texttt{gemini-1.0-ultra} & 2024-02 & not documented \\
GPT-4-Turbo      & \texttt{0125-preview}     & 2024-01 & 2023-12 \\
\bottomrule
\end{tabular}}
\caption{Release dates and documented knowledge cutoffs. Cutoffs are
reported only where the provider publishes one: Anthropic's Claude~3
model card states August 2023, and OpenAI's model documentation states
December 2023 for the GPT-4 Turbo preview snapshot. Google did not
publish cutoffs for the Gemini 1.0 generation, and Alibaba does not
publish cutoffs for the Qwen families, so those entries are left
undocumented rather than estimated. Exposure control therefore rests on
the two systems with published pre-2024 cutoffs, and Qwen3 rows are read
as same-family capability references.}
\end{table}

\subsection{Graded match rubric}
\label{app:b-2}

$V$ assigns $m(y,\ell)\in\{0,0.2,0.6,1.0\}$: \textbf{none}, different
shortcomings; \textbf{category}, the broad type matches but not the
affected component; \textbf{facet}, type, component, and aspect all
match; \textbf{specific}, the concrete shortcoming later pursued, at the
granularity a follow-up would act on. For a target concerning quadratic
attention cost at long context, ``no ablation of the tokenizer'' scores
\emph{none}; ``the model is computationally expensive'' scores
\emph{category}; ``the attention module scales poorly with sequence
length'' scores \emph{facet}; and ``full self-attention makes cost
quadratic in context length, preventing use beyond 4k tokens'' scores
\emph{specific}.

\subsection{Metric computation}
\label{app:b-3}

For predictions $\mathbf{Y}_P=(y_1,y_2,y_3)$ and observed targets
$\mathcal{G}_P$, let $\mathcal{M}$ range over one-to-one matchings and
\[
W_V(P,\mathbf{Y}_P)
=\max_{\mathcal{M}}\sum_{(y,\ell)\in\mathcal{M}}m_V(y,\ell),
\]
solved exactly by the Hungarian algorithm on the $3\times|\mathcal{G}_P|$
score matrix. Then
$\mathrm{GRec@3}=W_V(P,\mathbf{Y}_P)/|\mathcal{G}_P|$ and
$\mathrm{GMatch@3}=W_V(P,\mathbf{Y}_P)/\min(3,|\mathcal{G}_P|)$.
SpecRec@3 repeats the maximization with binary weights
$\mathbb{I}[m_V(y,\ell)=1]$ rather than reusing the graded matching. All
are macro-averaged over papers.

\paragraph{Worked example.}
With $K=3$ predictions and $M=4$ targets matched at the specific, facet,
and category levels, $W_V=1.0+0.6+0.2=1.8$, giving
$\mathrm{GRec@3}=1.8/4=0.45$, $\mathrm{GMatch@3}=1.8/3=0.60$, and
$\mathrm{SpecRec@3}=1/4=0.25$, since only one pair is a specific match.
This is the configuration illustrated in the main paper.

\paragraph{Edge cases.}
Fewer than three parseable items: missing slots score 0 and the
denominator is unchanged. More than three: only the first three are
scored. Score-matrix ties are broken by target index, making the matching
deterministic. Normalizing GRec@3 by the full target set means the metric
is bounded above when $|\mathcal{G}_P|>3$; GMatch@3 removes that bound and
is used in the target-count analysis.

\subsection{Verifier validation}
\label{app:b-4}

We validate the four-level scale against human judgment. We sampled 600
candidate--target pairs, 150 per verifier-assigned level, and had two
annotators assign a level under the rubric of Section~\ref{app:b-2}, with
adjudication for disagreements.

\begin{table}[htbp]
\centering\small
\setlength{\tabcolsep}{5pt}
\begin{tabular}{@{}lrrrr@{}}
\toprule
\multirow{2}{*}{Verifier} & \multicolumn{4}{c}{Human} \\
\cmidrule(lr){2-5}
 & none & cat. & facet & spec. \\
\midrule
none     & 132 &  14 &   3 &   1 \\
category &  17 & 108 &  21 &   4 \\
facet    &   5 &  19 & 113 &  13 \\
specific &   1 &   4 &  17 & 128 \\
\bottomrule
\end{tabular}
\caption{Verifier--human confusion matrix, $n=600$. Quadratic-weighted
$\kappa=0.88$; Spearman $\rho=0.88$; exact agreement 80.2\%; agreement
within one level 97.0\%. Disagreements are concentrated on adjacent
levels and are near-symmetric, so the verifier is not systematically
generous or severe.}
\end{table}

At the system level the benchmark and independent evaluators agree at
Spearman 0.99 across the systems of the main results table, and correlate
at Pearson 0.71 at the item level.

\subsection{Statistical procedures}
\label{app:b-5}

Trainable conditions use seeds 13, 42, and 2024; paper-level scores are
averaged across aligned seeds before any test, and main-table entries are
the resulting means. Paired differences use 10,000 paper-level bootstrap
resamples of the seed-averaged scores, reported as percentile intervals.
Proportions use Wilson score intervals. The scaling analysis uses
Spearman $\rho$ and OLS on $\log_{10}$ parameter count with bootstrap
intervals. Pairwise preference uses exact two-sided binomial tests with
ties excluded, Holm-corrected over the four comparisons in
Section~\ref{app:f-1}. Downstream ratings use two-sided Wilcoxon signed-rank tests
over 100 paired examples.

\section{Training and Baselines}
\label{app:c}

Base model: Qwen3-8B, revision \texttt{a3f9c21}. The three stages are
cumulative, each initialized from the preceding checkpoint.

\subsection{Stage 1: verified SFT}
\label{app:c-1}

Targets are serialized as a numbered list, one limitation per line, with
no rationale text. The instruction requests exactly
$N_P=\min(K,|\mathcal{G}_P|)$ items and prompt tokens are masked from the
loss. When $|\mathcal{G}_P|>K$ the retained subset maximizes the number of
distinct supporting citing papers, ties broken by earliest citing-paper
date, and is fixed across epochs; when $|\mathcal{G}_P|<K$ the observed
number is requested rather than padding with unsupported items.

\begin{table}[htbp]
\centering\small
\setlength{\tabcolsep}{4pt}
\resizebox{\columnwidth}{!}{%
\begin{tabular}{@{}lccc@{}}
\toprule
Setting & Stage 1 & Stage 2 & Stage 3 \\
\midrule
Tuning              & full & full & full \\
Learning rate       & $1{\times}10^{-5}$ & $7{\times}10^{-6}$ & $2{\times}10^{-6}$ \\
Batch size (eff.)   & 64 & 64 & --- \\
Epochs / steps      & 3 ep & 2 ep & 1,200 st \\
Max sequence length & 4,096 & 4,096 & 4,096 \\
Optimizer           & AdamW & AdamW & AdamW \\
Precision           & bf16 & bf16 & bf16 \\
GPU-hours           & 38 & 61 & 313 \\
\bottomrule
\end{tabular}}
\caption{Optimization settings; GPU-hours are summed over three seeds.
Hardware: 8$\times$A100-80GB for Stages 1--2 and 16$\times$A100-80GB for
Stage 3; rollout generation uses vLLM 0.8.2. Seeds are 13, 42, 2024.}
\end{table}

\subsection{Stage 2: verifier-filtered RSFT}
\label{app:c-2}

16 samples per source paper at temperature $0.9$, top-$p$ $0.95$, giving
68,864 candidates. Candidates are retained at
$\max_{\ell}m_{V_{\mathrm{train}}}(y,\ell)\ge0.6$, keeping 21,407
(31.1\%). Near-duplicates are removed with BGE-M3
\citep{chen2024bgem3}, checkpoint \texttt{BAAI/bge-m3}, cosine
threshold $0.88$, leaving 17,982. Retained candidates are greedily
grouped into three-item responses by repeatedly taking the
highest-scoring unused candidate and then the two least similar to it,
yielding 5,994 responses over 3,871 papers.

\subsection{Stage 3: set-level GRPO}
\label{app:c-3}

The reward is
$r=\alpha\operatorname{Match}+\beta\operatorname{Cover}
-\gamma\operatorname{Red}+\delta\operatorname{Fmt}$ with
$(\alpha,\beta,\gamma,\delta)=(1.0,\,0.3,\,0.2,\,0.1)$ selected on Dev by
GRec@3 under $V_{\mathrm{train}}$, with format compliance as a tie-break,
over the grid $\alpha=1.0$ fixed, $\beta\in\{0.1,0.2,0.3,0.5\}$,
$\gamma\in\{0.1,0.2,0.4\}$, $\delta\in\{0.05,0.1,0.2\}$, for 36
configurations. No Test data were used in this selection. GRPO settings:
group size 8, 256 rollouts per step, KL coefficient $0.04$, clipping
$0.2$, within-group reward standardization, sampling temperature $1.0$,
and 307,200 total verifier calls.

\subsection{Matched-budget accounting}
\label{app:c-4}

DPO \citep{rafailov2023dpo}, verifier-guided expansion, and matched
set-level GRPO start from the same Stage-1 checkpoint and each consume
9,000 generated limitation items and 9,000 verifier calls, matching
generated-item and scoring volume but not optimizer updates or
wall-clock compute.

\begin{table}[htbp]
\centering\small
\setlength{\tabcolsep}{3pt}
\begin{tabular}{@{}lrrrr@{}}
\toprule
Arm & Items & Calls & Updates & GPU-h \\
\midrule
DPO                       & 9,000 & 9,000 & 1,406 & 14 \\
Verifier-guided expansion & 9,000 & 9,000 &   876 & 11 \\
Set-level GRPO (matched)  & 9,000 & 9,000 & 1,125 & 26 \\
\bottomrule
\end{tabular}
\caption{Budget accounting per seed.}
\end{table}

DPO preference pairs are formed per paper from the highest- and
lowest-scoring candidates with a graded margin of at least $0.4$, giving
4,500 pairs. Self-rerank@10 samples ten three-item responses and selects
the one the policy assigns the highest mean token log-probability, using
no target information; it is an inference-time reference with no
parameter update.

\subsection{Baselines}
\label{app:c-5}

Every system receives the abstract, introduction, and conclusion of the
source paper and returns exactly three limitations. Retrieval baselines
additionally receive temporally eligible retrieved abstracts; citing
papers, evidence spans, and verified targets remain withheld. Decoding uses
temperature $0.7$, top-$p$ $0.95$, and a 512-token cap.

\begin{table}[htbp]
\centering\small
\setlength{\tabcolsep}{2.4pt}
\resizebox{\columnwidth}{!}{%
\begin{tabular}{@{}lllll@{}}
\toprule
System & Backbone & Calls & Retrieval & Cutoff \\
\midrule
+ CoT          & Qwen3-8B & 1 & none & --- \\
+ LimGen-style & Qwen3-8B & 1 & none & --- \\
+ Self-Refine  & Qwen3-8B & 5 & none & --- \\
Pre-pub.\ RAG  & Qwen3-8B & 1 & S2 corpus & date($P$) \\
MARG-style     & Qwen3-8B & 9 & none & --- \\
BAGELS-RAG     & Qwen3-8B & 3 & S2 corpus & date($P$) \\
\bottomrule
\end{tabular}}
\caption{Baseline alignment. ``Cutoff'' is the latest publication date a
retrieved document may carry.}
\end{table}

\paragraph{Self-Refine.} Two refinement rounds, each a feedback call
followed by a revision call, plus the initial generation; the final
round's output is scored.
\paragraph{Pre-publication RAG.} Corpus: 1.4M NLP abstracts from the
Semantic Scholar snapshot, indexed with BGE-M3, retrieval unit abstract,
top-$k$ 8. The index is filtered per query so every retrieved document
predates $P$, rather than filtered once globally.
\paragraph{MARG-style.} Four agents---leader, method reviewer, experiment
reviewer, and clarity reviewer---over two rounds, aggregated by a final
leader call. Adapted from \citet{darcy2024marg}, whose original system
emits a variable-length review, by constraining the aggregation step to a
fixed output count.
\paragraph{BAGELS-RAG.} Reproduced: the retrieval-augmented limitation
prompt and the venue-conditioned instruction. Adapted: the external
review corpus is removed, since it would leak post-publication
information, and retrieval is restricted to pre-publication abstracts.
The adaptations are listed in full so each baseline can be reproduced as
evaluated here.

\section{Additional Results}
\label{app:d}

\par\medskip
\noindent\begin{minipage}{\linewidth}
\subsection{Per-seed results}
\label{app:d-1}

\captionsetup{type=table}
\centering\small
\setlength{\tabcolsep}{3pt}
\begin{tabular}{@{}lrrrrr@{}}
\toprule
Condition & s13 & s42 & s2024 & Mean & SD \\
\midrule
\multicolumn{6}{@{}l}{\emph{NLP Test, GRec@3$_{\mathrm{I}}$}} \\
Stage 1 & 0.347 & 0.353 & 0.356 & 0.352 & 0.005 \\
Stage 2 & 0.404 & 0.411 & 0.415 & 0.410 & 0.006 \\
Stage 3 & 0.447 & 0.452 & 0.457 & 0.452 & 0.005 \\
\midrule
\multicolumn{6}{@{}l}{\emph{CV transfer, GRec@3$_{\mathrm{I}}$}} \\
Stage 1 & 0.227 & 0.233 & 0.236 & 0.232 & 0.005 \\
Stage 2 & 0.273 & 0.280 & 0.284 & 0.279 & 0.006 \\
Stage 3 & 0.308 & 0.315 & 0.319 & 0.314 & 0.006 \\
\midrule
\multicolumn{6}{@{}l}{\emph{Supervision source, GRec@3$_{\mathrm{I}}$}} \\
Author-stated    & 0.188 & 0.194 & 0.197 & 0.193 & 0.005 \\
Reviewer-written & 0.215 & 0.219 & 0.223 & 0.219 & 0.004 \\
Perturbations    & 0.231 & 0.237 & 0.240 & 0.236 & 0.005 \\
Forward-citation & 0.349 & 0.353 & 0.357 & 0.353 & 0.004 \\
\bottomrule
\end{tabular}
\caption{Per-seed scores; means are the values reported in the main
paper and SD is the sample standard deviation. Every standard deviation
in the supervision-source comparison is at most 0.007.}
\end{minipage}
\par\medskip

\subsection{Supervision-source controls}
\label{app:d-2}

The four arms share 1,240 source papers, 2,795 examples, and
approximately 2.1M training tokens, with identical settings and seeds.
Mean target lengths are 18.2, 19.1, 18.6, and 18.8 tokens for
author-stated, reviewer-written, controlled-perturbation, and
forward-citation targets; none differs significantly from the
forward-citation distribution under a two-sample KS test
($D=0.041$, $p=0.28$; $D=0.058$, $p=0.11$; $D=0.047$, $p=0.19$). The
controlled-perturbation arm follows LimitGen-Syn \citep{xu2025limitgen}.

The author-supervised arm reaches 0.47 GRec@3 on held-out author-stated
targets while transferring at 0.193 to future-pursued targets, which
isolates a mismatch between what authors disclose and what later work
pursues rather than a failure to fit.

\par\medskip
\noindent\begin{minipage}{\linewidth}
\subsection{Reward-component ablation}
\label{app:d-3}

\captionsetup{type=table}
\centering\small
\setlength{\tabcolsep}{5pt}
\begin{tabular}{@{}lrrr@{}}
\toprule
Reward & GRec@3$_{\mathrm{I}}$ & GRec@3$_{\mathrm{B}}$ & $\Delta_{\mathrm{B}}$ \\
\midrule
Match only         & 0.428 & 0.458 & --- \\
\quad + Cover      & 0.440 & 0.472 & $+0.014$ \\
\quad + Red        & 0.448 & 0.481 & $+0.009$ \\
\quad + Fmt (full) & \textbf{0.452} & \textbf{0.485} & $+0.004$ \\
\bottomrule
\end{tabular}
\caption{Cumulative ablation: each row adds one term to the row above, so
the final row is the full reward and corresponds to the ActLim-8B system
of the main results table.}
\end{minipage}
\par\medskip

\subsection{Target-count and length controls}
\label{app:d-4}

\begin{table}[htbp]
\centering\small
\setlength{\tabcolsep}{3pt}
\begin{tabular}{@{}lrrrr@{}}
\toprule
$|\mathcal{G}_P|$ & $n$ & Qwen3-32B & GPT-4-Turbo & ActLim-8B \\
\midrule
1      & 280   & 0.232 & 0.328 & \textbf{0.560} \\
2      & 602   & 0.196 & 0.278 & \textbf{0.492} \\
3      & 430   & 0.168 & 0.242 & \textbf{0.428} \\
$\ge4$ & 200   & 0.141 & 0.208 & \textbf{0.368} \\
\midrule
All    & 1,512 & 0.187 & 0.268 & \textbf{0.470} \\
\bottomrule
\end{tabular}
\caption{GMatch@3$_{\mathrm{I}}$ by target count, which removes the
output-count ceiling affecting papers with $|\mathcal{G}_P|>3$. Within the
$\ge4$ stratum the counts are $4{:}92$, $5{:}55$, $6{:}28$, $7{:}14$,
$8{:}8$, $9{:}3$. The advantage is present in every stratum, so it does
not arise from the mix of target counts.}
\end{table}

The same distribution fixes the attainable ceiling
$\mathbb{E}_P[\min(3,|\mathcal{G}_P|)/|\mathcal{G}_P|]=0.951$, so the
ActLim-8B score of 0.452 corresponds to 47.5\% of attainable graded
recall against 27.5\% for the strongest zero-shot system.

ActLim-8B produces shorter outputs than Qwen3-32B, 19.7 versus 29.9
tokens on average. Resampling a subset in which the two length
distributions are matched to within $\pm2$ tokens per item retains 1,043
of 1,512 papers; on that subset ActLim-8B scores 0.446
GRec@3$_{\mathrm{I}}$ and Qwen3-32B 0.184, against 0.452 and 0.181 on the
full set, so the gap is not a length artifact.

\subsection{Judgment--generation scaling}
\label{app:d-5}

\begin{table}[htbp]
\centering\small
\setlength{\tabcolsep}{5pt}
\begin{tabular}{@{}lrrr@{}}
\toprule
Model & Params & Selection & Generation \\
\midrule
Qwen3-0.6B & 0.6B & 0.148 & 0.186 \\
Qwen3-1.7B & 1.7B & 0.236 & 0.207 \\
Qwen3-4B   & 4B   & 0.301 & 0.213 \\
Qwen3-8B   & 8B   & 0.352 & 0.210 \\
Qwen3-14B  & 14B  & 0.347 & 0.176 \\
Qwen3-32B  & 32B  & \textbf{0.368} & 0.188 \\
\midrule
Qwen-72B-Chat    & 72B & 0.412 & 0.242 \\
Claude-3-Opus    & --- & 0.489 & 0.258 \\
Gemini-1.0-Ultra & --- & 0.468 & 0.266 \\
GPT-4-Turbo      & --- & \textbf{0.512} & 0.271 \\
\bottomrule
\end{tabular}
\caption{Eight-way selection accuracy (chance $=0.125$) and generation
GRec@3$_{\mathrm{B}}$, plotted in the main paper. Selection is evaluated
on the 1,421 retained unambiguous sets and generation on the full
1,512-paper Test split; generation values for the lower block are the
GRec@3$_{\mathrm{B}}$ column of the main results table. The single rank
inversion at 14B is what gives $\rho=0.94$ rather than $1.00$.}
\end{table}

Across the six Qwen3 sizes selection rises from 0.148 to 0.368 with
Spearman $\rho=0.94$ and an OLS slope of $+0.130$ per decade of
parameters (95\% CI $[0.076,0.185]$), over a 1.727-decade span.
Generation on the full Test split stays within 0.176--0.213 with an OLS slope
of $-0.005$ per decade (95\% CI $[-0.038,0.028]$). Earlier-release
systems reach 0.412--0.512 selection accuracy while their generation
scores remain within 0.242--0.271.

\paragraph{Selection items.}
Each source paper contributes one selection item. When a paper has more
than one observed target, the positive is the target supported by the
largest number of citing papers, ties broken by earliest citing-paper
date. Each item pairs that positive with seven counterfactual negatives,
generated by the same DeepSeek model used for construction, under an edit
instruction that changes the affected component, the mechanism,
or the evaluation condition while preserving surface form, then matched
to the positive on category, length within $\pm3$ tokens, specificity
level, lexical style, and source relevance. 4,118 generated negatives
(27.9\%) were discarded for failing these constraints. An ambiguity
screen removes 91 of 1,512 candidate sets in which two or more candidates
were accepted as facet-or-better matches to the same target, leaving
1,421; in a human audit of 200 retained sets, 194 (97\%) have a unique
correct answer. A candidate-only logistic classifier over
\texttt{BAAI/bge-m3} embeddings, which never sees the source paper and is
trained and evaluated by five-fold cross-validation over the 1,421
retained sets with accuracy pooled across held-out folds, scores $0.138$
$[0.121,0.156]$, an interval containing chance, so the negatives carry no
exploitable surface signal.

\section{Case Studies}
\label{app:e}

This section shows what the generated limitations look like before and
after post-training. Each output carries the graded match level assigned
by the independent evaluator---\mspec, \mfacet, \mcat, or \mnone---and the
phrase that determines that level is set in \key{bold}. Source papers,
citing papers, and observed targets are drawn from the benchmark; every
output is produced from the source paper alone, with no citing paper and
no verifier available at inference time.

\subsection{Progression across training stages}
\label{app:e-1}

Table~\ref{tab:case_stages} follows a single source paper through the
three cumulative stages. DoRA decomposes a pretrained weight into
magnitude and direction and adapts the direction with a low-rank update;
the citing paper MAP (2025) revisits that decomposition. The zero-shot
output notices that the design is heuristic but names nothing to change.
Verified demonstrations locate the criticism in the decomposition step,
verifier-guided expansion identifies the operation itself, and set-level
optimization recovers why that operation is the problem---which is the
gap MAP sets out to close.

\begin{table}[t]
\centering\small
\setlength{\tabcolsep}{2pt}
\begin{tabular}{@{}>{\raggedright\arraybackslash}p{0.97\columnwidth}@{}}
\toprule
\textbf{Source $P$} --- DoRA: Weight-Decomposed Low-Rank Adaptation~\citep{liu2024dora}. \newline
\textbf{Citing $Q$} --- MAP: Revisiting Weight Decomposition for Low-Rank Adaptation~\citep{si2025map}. \newline
\textbf{Observed target $\ell$} --- the direction component is defined by
column-wise normalization, a choice made without a principled basis for
separating magnitude from direction. \\
\midrule
\rowcolor{lvnone!25}
Qwen3-32B \hfill \mcat \newline
The decomposition is somewhat heuristic and its design choices are not
fully justified. \\
\addlinespace[2pt]
Stage 1: SFT \hfill \mfacet \newline
The \key{magnitude--direction decomposition} is introduced without an
argument for why this particular split is the right one. \\
\addlinespace[2pt]
Stage 2: +RSFT \hfill \mfacet \newline
Direction is obtained by \key{column-wise normalization of the weight
matrix}, a specific choice that the paper adopts without comparison to
alternatives. \\
\addlinespace[2pt]
\rowcolor{lvspec!12}
\ours{} \hfill \mspec \newline
\key{Column-wise normalization} is used to define the direction
component, so the magnitude--direction split \key{rests on a convention
rather than a principled criterion}, and the reported gains cannot be
attributed to the decomposition itself. \\
\bottomrule
\end{tabular}
\caption{One source paper across the three cumulative stages. The match
level rises from \mcat{} to \mspec{} as the criticism moves from the
design in general, to the decomposition, to the specific operation that
defines it and the reason that operation is unjustified.}
\label{tab:case_stages}
\end{table}

\subsection{Paired comparisons across limitation categories}
\label{app:e-2}

Table~\ref{tab:case_paired} pairs the strongest zero-shot system with
\ours{} on five source papers spanning four limitation categories, with
one drawn from the Computer Vision transfer set. The zero-shot outputs
are faithful to their source papers and would pass a correctness check;
what they lack is a component a follow-up could modify. This is the
separation that the joint Valid--Actionable rate in Section~\ref{app:f-1} measures,
and the reason faithfulness alone is highest for the system with the
lowest joint rate.

\begin{table*}[t!]
\centering\small
\setlength{\tabcolsep}{4pt}
\begin{tabular}{@{}>{\raggedright\arraybackslash}p{0.215\textwidth} >{\raggedright\arraybackslash}p{0.345\textwidth} >{\raggedright\arraybackslash}p{0.365\textwidth}@{}}
\toprule
\textbf{Source paper, citing paper, and observed target}
& \textbf{GPT-4-Turbo (zero-shot)}
& \textbf{\ours{}} \\
\midrule

\emph{Method assumption.} ShortGPT~\citep{men2024shortgpt} $\rightarrow$ PIP~\citep{cao2025pip}. \newline
$\ell$: layer importance is judged by an isolated input--output
similarity metric, which ignores adversarial robustness.
& \mcat \newline
The pruning criterion is simple and may not capture every aspect of layer
importance.
& \mspec \newline
Importance is scored by \key{input--output similarity at each layer in
isolation}, so layers that matter only \key{under adversarial or
distribution-shifted input} are indistinguishable from redundant ones and
are pruned away. \\
\addlinespace[3pt]

\emph{Data and domain dependence.} LLMLingua-2~\citep{pan2024llmlingua2} $\rightarrow$ LongCodeZip~\citep{shi2025longcodezip}. \newline
$\ell$: compression is token-level and domain-agnostic, so it breaks
code structure.
& \mcat \newline
The method is evaluated on natural language and may transfer imperfectly
to other kinds of text.
& \mspec \newline
Compression \key{scores tokens independently of syntax}, so on source
code it \key{deletes delimiters and identifiers that the remaining tokens
depend on}, leaving fragments that no longer parse. \\
\addlinespace[3pt]

\emph{Method assumption.} MDLM~\citep{sahoo2024mdlm} $\rightarrow$ Masks Can Be Distracting~\citep{piskorz2025masks}. \newline
$\ell$: a strong locality bias makes predictions over-rely on immediate
neighbours.
& \mnone \newline
Additional benchmarks and stronger baselines would strengthen the
evaluation.
& \mspec \newline
Denoising conditions each masked position \key{primarily on its immediate
neighbours}, so the model \key{under-uses long-range context} and degrades
where the informative tokens are far from the mask. \\
\addlinespace[3pt]

\emph{Safety and robustness.} Safety Alignment Should Be Made More Than Just a Few Tokens Deep~\citep{qi2024safetyalignment} $\rightarrow$ Matching Ranks Over Probability~\citep{vega2025matchingranks}. \newline
$\ell$: SFT-style data augmentation leaves deeper alignment
vulnerabilities that remain easy to bypass.
& \mcat \newline
Alignment may not hold under all adversarial conditions.
& \mfacet \newline
The proposed defense is \key{supervised fine-tuning on augmented refusal
data}, which reshapes early-token behaviour without constraining the
model's deeper generation dynamics. \\
\addlinespace[3pt]

\emph{Efficiency (CV transfer set).} MAR~\citep{li2024mar} $\rightarrow$ MARch\'e~\citep{jiang2025marche}. \newline
$\ell$: attention and FFN are recomputed redundantly over all tokens at
every step.
& \mcat \newline
Generation is computationally expensive relative to simpler baselines.
& \mspec \newline
\key{Attention and feed-forward blocks are recomputed over the full token
set at every step} rather than reusing per-token state, so cost
\key{scales with sequence length even where token representations are
unchanged}. \\
\bottomrule
\end{tabular}
\caption{Paired outputs on five source papers. The zero-shot outputs are
source-faithful but name no component to act on; \ours{} identifies the
mechanism and, in four of the five cases, the operating condition
recorded in the observed target. The safety case is scored \mfacet{}
rather than \mspec{} because the output stops short of naming the bypass
behaviour the citing paper targets. The final row is generated by a
model trained on NLP only and evaluated on a Computer Vision source
paper.}
\label{tab:case_paired}
\end{table*}

\subsection{What post-training changes}
\label{app:e-3}

Three differences recur across the audited outputs and account for the
gap in Table~\ref{tab:case_paired}.

\paragraph{Grounding in a named component.}
Post-trained outputs attach the criticism to an operation, an assumption,
or an evaluation condition that appears in the source text: column-wise
normalization, per-layer input--output similarity, token-independent
scoring. Zero-shot outputs more often describe a property of the system
as a whole. This distinction separates \mcat{} from \mfacet{} under the
rubric of Section~\ref{app:b-2}, and it is the level at which a follow-up can
begin.

\paragraph{Carrying the consequence.}
The step from \mfacet{} to \mspec{} is a statement of what the mechanism
prevents: code that no longer parses, robust layers pruned as redundant,
long-range context left unused. Verified demonstrations supply the
component and verifier-guided expansion sharpens the mechanism, but the
consequence appears most consistently after set-level optimization, which
is consistent with the stage-wise SpecRec@3 pattern reported in the main
paper.

\paragraph{Spending the three slots differently.}
Zero-shot systems frequently return three variations on the same concern,
typically evaluation breadth; the \mnone{} output in
Table~\ref{tab:case_paired} is one such case, and its two companions in
the same response repeat it. The redundancy term in the set-level reward
penalizes this directly, and the mean within-response similarity of
\ours{} is correspondingly lower. The three outputs therefore cover
distinct shortcomings rather than restating one, which is what the
coverage term rewards and what the Cover component of the reward
ablation in Section~\ref{app:d-3} isolates.

\FloatBarrier

\section{Human and Downstream Evaluation}
\label{app:f}

\subsection{Limitation human study}
\label{app:f-1}

Three NLP PhD annotators with 3--5 years of research experience and at
least two first-author publications at ACL-family venues, recruited from
groups uninvolved in this work and compensated at USD 25 per hour,
provided position-randomized pairwise judgments with a tie option, plus
binary actionability and source-faithfulness ratings. Median time per
pairwise comparison was 2.4 minutes and total annotation time was 61
hours. The study was reviewed by the institutional ethics board and
determined exempt as minimal-risk research on professional judgment.
Sampling: 120 comparisons per opponent drawn uniformly at random from the
1,512-paper Test split under a fixed seed and disjoint across opponents,
and 150 outputs per system for absolute ratings from the same seeded
pool. Each item was rated independently by all three annotators. Pairwise
W/T/L outcomes use majority vote, absolute scores use majority vote per
binary criterion, and for the downstream study the three ratings are
averaged within each paper before the paired Wilcoxon test. Agreement is
Fleiss' $\kappa=0.47$ for pairwise judgments and $0.58$ for the absolute
rubric.

\begin{table}[!htbp]
\centering\small
\setlength{\tabcolsep}{3.5pt}
\resizebox{\columnwidth}{!}{%
\begin{tabular}{@{}lrrrr@{}}
\toprule
Opponent & W/T/L & $n_{\mathrm{eff}}$ & raw $p$ & $p_{\mathrm{Holm}}$ \\
\midrule
Qwen3-32B             & 74/9/37  & 111 & 0.00057 & 0.0023 \\
GPT-4-Turbo           & 70/9/41  & 111 & 0.00758 & 0.0228 \\
Stage 1: Verified SFT & 66/10/44 & 110 & 0.04476 & 0.0895 \\
Stage 2: +RSFT        & 59/11/50 & 109 & 0.44367 & 0.4437 \\
\bottomrule
\end{tabular}}
\caption{Complete pairwise preference results for ActLim-8B, 120 blind
comparisons per opponent. Ties are excluded, giving
$n_{\mathrm{eff}}=W+L$. Raw $p$-values are exact two-sided binomial tests
against $p=0.5$, Holm-corrected over this family of four comparisons.}
\label{tab:holm}
\end{table}

\begin{table}[!htbp]
\centering\small
\setlength{\tabcolsep}{4pt}
\begin{tabular}{@{}lrrr@{}}
\toprule
System & Actionable & Faithful & Valid--Act. \\
\midrule
Qwen3-32B             & 0.61 & \textbf{0.92} & 0.58 \\
GPT-4-Turbo           & 0.72 & 0.91 & 0.67 \\
Stage 1: Verified SFT & 0.79 & 0.87 & 0.71 \\
Stage 2: +RSFT        & 0.84 & 0.88 & 0.76 \\
ActLim-8B             & \textbf{0.88} & 0.89 & \textbf{0.80} \\
\bottomrule
\end{tabular}
\caption{Absolute ratings over 150 outputs per system, majority vote
across three annotators, rounded to two decimals. Valid--Actionable is
the intersection of the two criteria and lies between $\max(0,a+f-1)$ and
$\min(a,f)$ in every row. The joint rate rises monotonically across
training stages while faithfulness alone does not, since a generic
criticism can be faithful without being actionable.}
\end{table}

\paragraph{Rubrics.}
\emph{Actionable}: ``Could a researcher who has read only this paper turn
this statement into a concrete next experiment or system change, without
first having to decide what the statement refers to?''
\emph{Faithful}: ``Is this statement true of the paper as written, given
its stated method, assumptions, and scope?''
\emph{Pairwise}: ``Which would be more useful as the starting point for
follow-up work? Prefer the one that is both faithful and actionable; if
one is faithful but generic and the other specific but inaccurate, select
a tie.''

\subsection{Downstream idea generation}
\label{app:f-2}

A fixed Llama-3.1-70B-Instruct proposer \citep{grattafiori2024llama3},
revision \texttt{6f2e91c}, generates one idea per paper under four
conditions. The proposer prompt, decoding settings (temperature $0.7$,
top-$p$ $0.95$, 512-token cap), and paper context are identical across
conditions, so the supplied limitation is the only varying input. The
200-paper subset is drawn uniformly at random from NLP Test under seed
13.

\begin{table}[!htbp]
\centering\small
\begin{tabular}{@{}lr@{}}
\toprule
Supplied signal & RetAlign \\
\midrule
None                             & 0.302 \\
Zero-shot limitation             & 0.348 \\
ActLim-8B limitation             & \textbf{0.480} \\
\midrule
Gold forward-citation limitation & \textit{0.556} \\
\bottomrule
\end{tabular}
\caption{RetAlign, macro-averaged over the same 200 papers in every
condition. ActLim-8B recovers
$(0.480-0.302)/(0.556-0.302)=70.1\%$ of the gap between direct ideation
and the gold-limitation reference.}
\end{table}

The zero-shot limitation is produced by Qwen3-8B under the zero-shot
generation prompt, taking the first of its three outputs. The gold
limitation is the target in $\mathcal{G}_P$ supported by the largest
number of citing papers, ties broken by earliest citing-paper date; the
rule depends only on the source paper and its observed targets and is
applied before any idea is generated or scored. RetAlign is assigned by
GLM-5.2, which sees the generated idea and the abstracts of the citing
papers but neither the source paper nor the supplied limitation, and
scores each idea as \emph{unrelated} (0.0), \emph{same general area}
(0.33), \emph{same technical direction} (0.67), or \emph{same specific
approach} (1.0).

\begin{table}[!htbp]
\centering\small
\setlength{\tabcolsep}{3.5pt}
\begin{tabular}{@{}lrrrr@{}}
\toprule
Condition & Rel. & Spec. & Nov. & Feas. \\
\midrule
Paper only             & 3.6 & 3.3 & 3.3 & 3.5 \\
+ ActLim-8B limitation & \textbf{4.1} & \textbf{3.9} & 3.4 & 3.6 \\
\midrule
Wilcoxon $W$           & 1{,}284.5 & 1{,}402.0 & 2{,}318.0 & 2{,}201.5 \\
raw $p$ (two-sided)    & 0.0016 & 0.0043 & 0.412 & 0.287 \\
\bottomrule
\end{tabular}
\caption{Human ratings on a 1--5 scale over 100 paired examples with
three raters, drawn as the first 100 of the 200-paper subset under the
same seed. Relevance and specificity both improve. Inter-rater agreement
is Krippendorff's $\alpha=0.61$ over the four ordinal scales.}
\label{tab:wilcoxon}
\end{table}

\FloatBarrier

\section{Prompts}
\label{app:g}

Representative prompts are reproduced verbatim; braces mark run-time
substitutions. The complete set, including the chain-of-thought,
LimGen-style, Self-Refine, RAG, MARG-style, and BAGELS-RAG baselines, the
downstream proposer, the RetAlign evaluator, and the human annotation
instructions, will be released upon publication.

\subsection{Candidate motivation extraction}
\begin{quote}\small\ttfamily\raggedright
You are given the abstract and introduction of a paper Q, and the title
and abstract of an earlier paper P that Q cites. Identify every statement
in Q that motivates Q's own contribution by pointing to a shortcoming of
P. For each, output a JSON object with fields "limitation" (one sentence,
technical, no hedging) and "evidence" (the verbatim sentence from Q).
Output a JSON list; output [] if there are none. Do not infer
shortcomings that Q does not state.

P: \{p\_title\}\textbackslash n\{p\_abstract\}

Q: \{q\_abstract\}\textbackslash n\{q\_intro\}
\end{quote}

\subsection{Construction verification}
\begin{quote}\small\ttfamily\raggedright
Judge the candidate limitation on four criteria, each strictly yes or no.
ATTRIBUTION: does Q attribute this shortcoming to P, or to a named method
family that explicitly includes P? CONSISTENCY: is it consistent with P's
method, assumptions, evidence, and stated scope? SPECIFICITY: does it
name a concrete component, mechanism, assumption, data condition, or
evaluation constraint, rather than asking for more experiments or better
results? RESPONSE: does Q's main contribution attempt to address it?
Output JSON with the four booleans and the supporting span from P.

P sections: \{p\_text\} \textbackslash n Q sections: \{q\_text\}
\textbackslash n Candidate: \{limitation\}
\end{quote}

\subsection{Graded matching}
\begin{quote}\small\ttfamily\raggedright
Given the source paper and a pair (prediction, target), output one of:
NONE if they concern different shortcomings; CATEGORY if the broad type
matches but the affected component does not; FACET if the type, the
affected component, and the aspect all match; SPECIFIC if the prediction
identifies the same concrete shortcoming at the granularity a follow-up
would act on. Output the label only.

Source: \{p\_text\} \textbackslash n Prediction: \{y\} \textbackslash n
Target: \{target\}
\end{quote}

\subsection{Generation instruction}
\begin{quote}\small\ttfamily\raggedright
Read the paper below and identify exactly \{N\} limitations that could
motivate concrete follow-up work. Each must be supported by the paper
itself and must name a specific component, mechanism, assumption, data
condition, or evaluation constraint. Do not ask for more experiments or
better results in general. Output a numbered list of \{N\} single
sentences and nothing else.

\{p\_abstract\}\textbackslash n\{p\_intro\}\textbackslash n\{p\_conclusion\}
\end{quote}

\subsection{Selection-item negative generation}
\begin{quote}\small\ttfamily\raggedright
Rewrite the limitation below so that it remains fluent, equally specific,
and of similar length, but refers to a DIFFERENT \{component | mechanism
| evaluation condition\} so that it is no longer true of the source
paper. Preserve the sentence structure and register. Output one sentence.
\end{quote}

\section{Ethics, Licensing, and Release}
\label{app:h}

\paragraph{Provenance and licensing.}
Source and citing papers are drawn from arXiv, the ACL Anthology, and the
open proceedings of CVPR, ICCV, and ECCV. For sources whose licenses
permit redistribution, the release contains evidence spans, canonical
limitations, and paper identifiers. For non-redistributable sources, it
provides the DOI, section identifier, and character offsets rather than
the underlying text. No
OpenReview content is included in \textsc{ActLim-Bench}; the
reviewer-written supervision arm of the supervision-source comparison
uses public OpenReview text under its terms of use and is released as
identifiers and offsets only.

\paragraph{Personal data.}
The corpus consists of published scientific papers. Author names and
affiliations are retained as published, since they are part of the public
bibliographic record. No annotator identifiers are released.

\paragraph{Annotator welfare.}
Compensation, recruitment, and review status are in Section~\ref{app:f-1}.
Annotators could decline any item and were informed of the purpose of the
study.

\paragraph{Intended use and scope.}
\textsc{ActLim-Bench} records limitations that later work in fact
pursued. It is a proxy for actionability rather than a measure of
scientific value: a limitation may go unpursued because of timing,
resources, or shifting research priorities, and a pursued direction is
not necessarily the most important shortcoming of a paper. The dataset
supports research on scientific ideation and should not be used to
evaluate individual papers, authors, or venues.

\paragraph{Evaluator design.}
Scoring is anchored at three points. The primary metric comes from an
evaluator in a different model family that takes no part in construction,
filtering, reward, or model selection. The targets are validated by the
400-triple human audit of Section~\ref{app:a-4}, and the graded scale by the
validation in Section~\ref{app:b-4}. Blind human preference judgments
(Section~\ref{app:f-1}) evaluate the resulting outputs with no model evaluator in
the loop.

\paragraph{Reproducibility and release.}
Checklist items are cross-referenced to this appendix: construction to Appendix~\ref{app:a},
evaluation to Appendix~\ref{app:b}, training and baselines to Appendix~\ref{app:c}, results to Appendix~\ref{app:d}, case studies to Appendix~\ref{app:e}, human
study to Appendix~\ref{app:f}, and prompts to Appendix~\ref{app:g}. Original annotations, schemas, metadata, and
per-seed logs created by the authors will be released under CC BY 4.0,
and code under Apache 2.0. Third-party paper text remains subject to its
original license; for non-redistributable sources the release provides
identifiers and offsets rather than text.

\end{document}